\documentclass[11pt]{article}

\usepackage{amssymb}
\usepackage{amsthm}
\usepackage{amsmath}
\usepackage{linguex}
\usepackage{booktabs}
\usepackage{multirow}
\usepackage{graphicx}
\usepackage{siunitx}
\usepackage{caption}
\usepackage{subcaption}
\usepackage{times}
\usepackage{inconsolata}
\usepackage{colortbl}
\usepackage{soul}
\usepackage{natbib}
\usepackage{xurl}
\usepackage{epigraph}
\usepackage{wrapfig}

\usepackage[margin=1in]{geometry}
\usepackage[colorlinks = true,
           linkcolor = MidnightBlue,
           urlcolor  = MidnightBlue,
           citecolor = MidnightBlue,
           anchorcolor = MidnightBlue]{hyperref}

\newcommand{\revision}[1]{#1}

\usepackage{utils}

\newcommand{\haap}{\textsc{haap}}

\newcommand{\haapmax}{\textsc{haap}$_{\textit{max}}$}
\newcommand{\haapmin}{\textsc{haap}$_{\textit{min}}$}

\usepackage[dvipsnames]{xcolor}

\DeclareSymbolFontAlphabet{\mathnormal}{letters}
\DeclareSymbolFont{letters}{OML}{cmm}{m}{it}
\DeclareSymbolFont{symbols}{OMS}{cmsy}{m}{it}
\DeclareMathAlphabet\mathcal{OMS}{cmsy}{m}{it}
\SetMathAlphabet\mathcal{bold}{OMS}{cmsy}{b}{it}

\newcommand{\naba}{\textsc{naba}}
\newcommand{\nabas}{\textsc{naba}s}
\newcommand{\nana}{\textsc{nana}}
\newcommand{\nanas}{\textsc{nana}s}

\newcommand{\blank}{$\rule{0.6cm}{0.15mm}$}

\usepackage{lineno}

\title{A systematic framework for generating novel experimental hypotheses from language models}

\date{}
\author{
\textbf{Kanishka Misra$^{*, 1}$}\\
Department of Linguistics\\
The University of Texas at Austin\\
\texttt{kmisra@utexas.edu}
\and 
\textbf{Najoung Kim$^{*, 2}$}\\
Department of Linguistics\\
Boston University\\
\texttt{najoung@bu.edu}
}

\begin{document}
\maketitle


\epigraph{\textit{They then took this machine and they ``turned the handle on the crank'' and said, ``let's see what else it will do'', and it turned out to generate a bunch of behaviors that also were not obvious, but which they then proceeded to test with human subjects...and lo and behold the human subjects, to everybody's surprise, acted just like the machine.}}{\footnotesize Jeff Elman on \citet{rumelhart1982interactive} during an Interview with Roger Bingham at CogSci 2010}

\begin{abstract}
\noindent
Neural language models (LMs) have been shown to capture complex linguistic patterns, yet their utility in understanding human language and more broadly, human cognition, remains debated. While existing work in this area often evaluates human-machine alignment, few studies attempt to translate findings from this enterprise into novel insights about humans. To this end, we propose a systematic framework for hypothesis generation that uses LMs to simulate outcomes of experiments that do not yet exist in the literature. We instantiate this framework in the context of a specific research question in child language development: dative verb acquisition and cross-structural generalization. Through this instantiation, we derive two novel, untested hypotheses: (1) the alignment between argument ordering and discourse prominence features of exposure contexts modulates how children generalize new verbs to unobserved structures; and \revision{(2) that this alignment in the first postverbal argument of the exposure context has a more noteworthy effect.} Additionally, we also design a set of experiments that can test these hypotheses in the lab with children. This work contributes both a domain-general framework for systematic hypothesis generation via simulated learners and domain-specific, lab-testable hypotheses for child language acquisition research.
\end{abstract}


\noindent
\section{Introduction}
\label{sec:intro}

Recent advances in Artificial Intelligence powered by neural network language models (LMs)\footnote{By neural network LMs, we mean any neural network model that is trained using a predictive objective (e.g., next-token prediction or masked language modeling objective) on a corpus of sequences in a given language.} trained at scale have generated a series of discussions about their role in (human) cognitive science \citep[\textit{i.a.}]{ambridge2020abstractions, toneva2021bridging, piantadosi2023modern,kodner2023linguistics, mcgrath2023can, portelance-jasbi-2024-roles, futrell2025linguistics}. A significant portion of such discussions are not limited to language modeling as a training objective or the dominant model architecture per se, and are more broadly applicable to the role of artificial neural networks (ANNs) \citep{pater2019generative,cichy2019deep,baroni2020linguistic,warstadt2022artificial}. Overall, both the arguments for and against the utility of LMs or ANNs are heavily reminiscent of debates from the second wave of connectionism \citep[\textit{i.a.}]{mcclelland1988connectionist,massaro1988some,fodor1988connectionism,smolensky1991constituent,mccloskey1991networks,hadley1997cognition}. Questions about whether connectionist models/ANNs/LMs count as theory, whether their successful replication of human behavior has any implications for advancing our understanding of human cognition, and discussion about their bearings on learnability arguments, are some recurring themes in these debates. While there are disagreements about the utility of contemporary LMs for cognitive investigations along these dimensions and more, one general consensus is that cautions must be taken in accepting them directly as models \textit{of} human cognition \citep{guest2023logical}, despite their strong predictive capabilities of behavioral \citep[][\textit{i.a.}]{goodkind-bicknell-2018-predictive, wilcox2020predictive, shain2024large} and neural data \citep[][\textit{i.a.}]{schrimpf2021neural, goldstein2022shared}. This is mainly due to sparse linking hypotheses between components of the human cognitive capacity and components of widely adopted models such as architecture, data, training process, and input/output representations, leading to limitations in explanations that can be offered by studies of these models alone.

Nevertheless, following \citet{mccloskey1991networks}, we argue that treating these models as \textit{animal models} is a promising way that black box models with high predictive capacity can contribute to cognitive science, and empirically explore this possibility. Specifically, we provide a concrete case study where we use LMs as simulated learners to derive novel experimental hypotheses that can in turn be tested with actual human learners. We expect this approach to be the most fruitful for domains in which large-scale human experiments are challenging, such as child language acquisition. \revision{There are existing human experimental findings implicated by neural network studies (e.g., predictions of \citealt{portelance-etal-2021-emergence} being corroborated by a concurrent human study of \citealt{jara2022origins}), and also findings that have been motivated by neural network studies (e.g., findings of \citealt{kim-linzen-2020-cogs} motivating the human experiments of \citealt{kim2024structural}). However, limited work has been \textit{explicitly} designed for the goal of novel hypothesis generation in the experimental linguistics space (with \citet{lakretz2021mechanisms}'s study on agreement in adults being a rare exception).} Furthermore, no such studies to the best of our knowledge exist in the language acquisition literature. This work attempts a proof-of-concept implementation of this idea using the acquisition of dative alternation as a case study. 

\begin{figure}
    \centering
    \includegraphics[width=\linewidth]{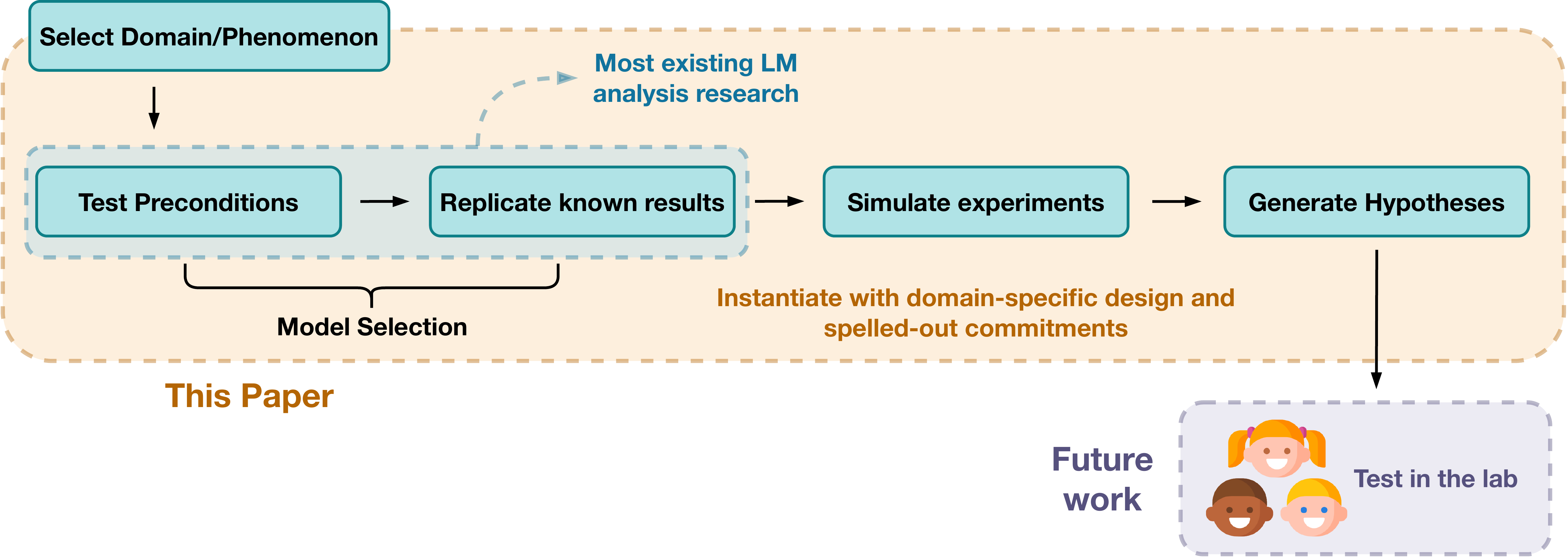}
    \caption{Overview of our broader pipeline for hypothesis generation from language models.}
    \label{fig:pipeline}
\end{figure}

If we were to take the idea of ``hypothesis generation'' seriously, an evident gap in the literature is the lack of a systematic framework that allows us to move beyond opportunistic discoveries of hypotheses during model analysis. Therefore, our starting point is proposing a more general framework of hypothesis generation that consists of five steps (Figure~\ref{fig:pipeline}): Domain selection, Precondition testing, Replication of known experimental results, Simluation, and Hypothesis generation. We view this general framework also as a major contribution of our work, in addition to the actual empirical study. Below, we illustrate how the general framework and its pipeline components get instantiated with respect to a specific problem domain.

\subsection{Domain-specific Instantiation of the Hypothesis Generation Pipeline: Acquisition of Dative Alternation}
\label{subsec:intro-case-study}

\paragraph{Step 0: Select the problem domain} Our problem domain is children's acquisition of the dative alternation. The dative alternation in English refers to a phenomenon where both prepositional phrase (PO: \ref{ex:dative-alternation-pp}) and double object (DO: \ref{ex:dative-alternation-do}) constructions are licensed for the same verb (in the example, for \textit{give}). One important difference between the two constructions is the linear ordering of the theme and recipient arguments: in PO, the theme precedes the recipient, and in DO, the recipient precedes the theme. 

\ex. \label{ex:dative-alternation}
    \a. \label{ex:dative-alternation-pp} Mikaylah gave a treat to Luna. (a treat = theme, Luna = recipient)
    \b. \label{ex:dative-alternation-do} Mikaylah gave Luna a treat. (Luna = recipient, a treat = theme)

\noindent However, not all dative verbs participate in this alternation; some only appear to be licensed in either the PO or the DO construction (*\textit{I donated the library the book}, *\textit{He wished luck to me}). Hence, learning which constructions a dative verb is licensed in is a problem that language learners face. It has been shown empirically that learners often \textit{generalize} to the alternate construction even in the absence of direct evidence showing that the verb is licensed in the alternate construction (e.g., \citet{gropen1989learnability,arunachalam2017preschoolers}), rather than being fully conservative and restricting the use of the dative verb to only the observed construction. Then, how does a learner distinguish non-alternating verbs from alternating verbs that they happened to not have observed positive evidence for one of the licensed constructions?
Two explanations are possible: first, there may be non-arbitrary \textit{criteria} that can tease these cases apart (the criteria hypothesis: \citet{gropen1989learnability}), and second, learners may have access to negative evidence that blocks overgeneralization. In this work, we focus on the first possibility and explore the criteria-based resolution to this learning problem, which falls under a broader class of learning problem known as Baker's paradox \citep{baker1979syntactic}.

A large body of prior work explores possible criteria that set apart alternating and non-alternating verbs, ranging from morphophonological to semantic factors (see \citet{citko2017double} for an overview). Many such factors are identified on the basis of distributional evidence (how often do we observe a dative verb with X/Y/Z properties in DO and PO constructions in a large corpus?) and adult acceptability judgments (is a dative verb with X/Y/Z properties acceptable in DO and PO constructions?). These studies are important in understanding which cues are available to the learners, since statistical cues in the input critically shape language acquisition \citep{saffran1996statistical, thompson2007statistical, romberg2010statistical}. However, the mere \textit{availability} of distinctive cues does not necessarily entail that they are \textit{used} by the learners. In this regard, nonce word studies testing whether learners generalize in the absence of observing the verb in the alternate form provide the most direct answer for the causal question: What cues do learners actually use to distinguish alternating verbs from non-alternating verbs? In adult learners, \citet{gropen1989learnability} identified factors such as possessive semantics and the number of syllables to affect PO to DO generalization, and \citet{coppock2009logical} found an effect of number of syllables as well as null results on the effect of prosodic weight and etymology (Germanic vs. Latinate).

In children, empirical evidence is sparser; only a handful of studies have investigated generalization to the alternate construction \citep{gropen1989learnability,conwell2007early,arunachalam2017preschoolers}, in addition to several studies that investigate the comprehension of dative structures \citep{rowland2010role,conwell2019effects} that may speak to the preconditions to generalization.  The sparse empirical evidence in children's acquisition of the dative alternation is partly due to the large size of the hypothesis space (combination of a wide range of distributional cues available in the input, as identified in the literature) and the difficulty in recruiting target participants at scale. To this end, we propose to use LMs as simulated learners to systematically explore this hypothesis space to identify a small number of targeted hypotheses, which in turn can be tested with actual children. We specifically investigate the role of distributional cues (as opposed to cues from the form of the verb itself) in the LM learners' generalization of novel dative verbs encountered in only one of the alternate constructions. Features of the context that the verb appears in, such as theme/recipient animacy, definiteness and length, have been shown to predict the choice between DO or PO in adult and child production \citep{bresnan2007predicting,de2012statistical} and have informed design decisions of nonce verb learning studies \citep{arunachalam2017preschoolers}. However, little is known about how these distributional cues get used by the learners to shape their inference about the alternation pattern of new verbs being learned.

\begin{wrapfigure}{R}{0.42\textwidth}
\vspace{-3.5em}
  \begin{center}
    \includegraphics[width=0.4\textwidth]{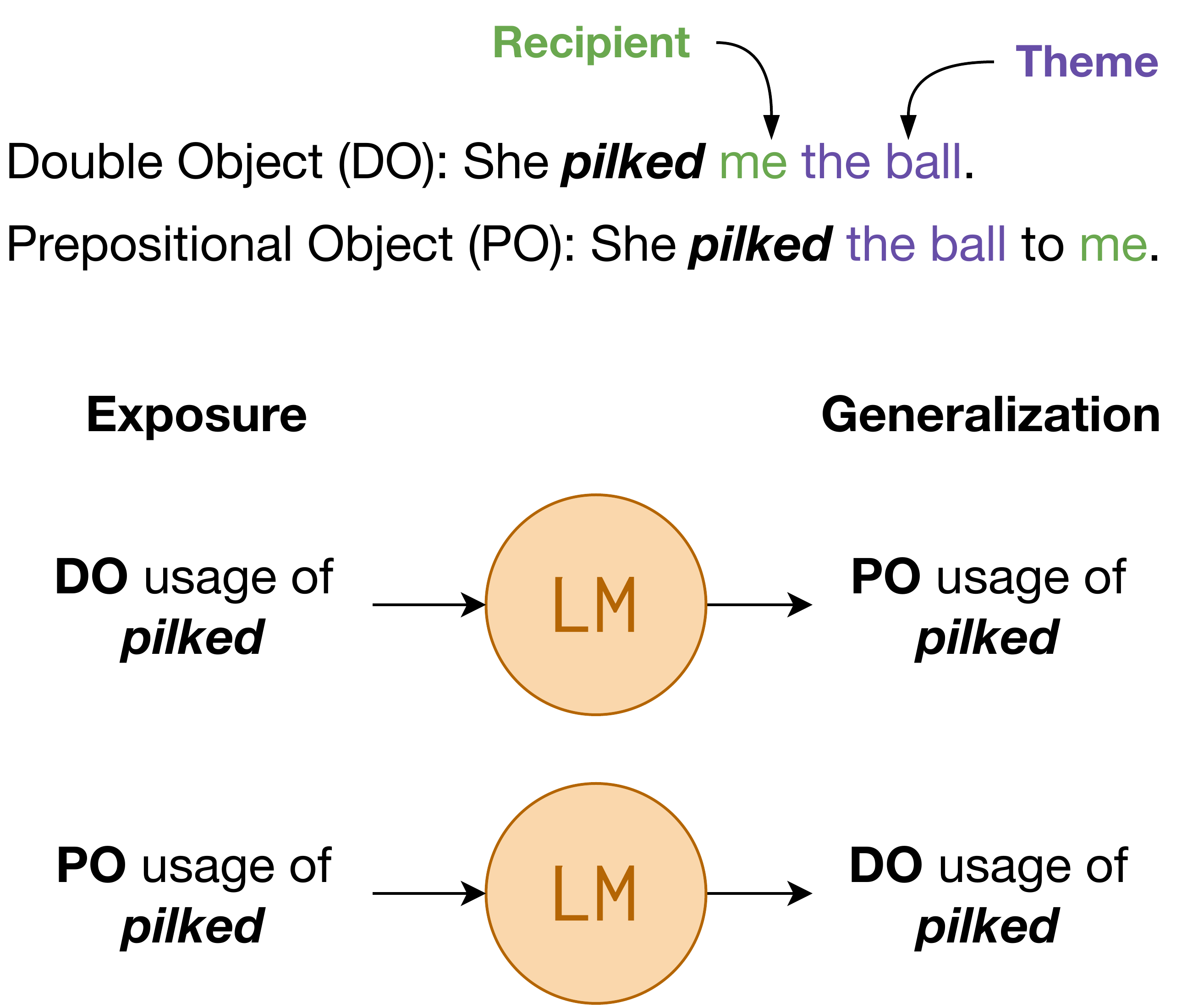}
  \end{center}
  \vspace{-1em}
   \caption{Depiction of cross-dative generalization, when the learner is exposed to a novel verb in one dative construction (exposure) and evaluated on a stimulus where the verb is used in the alternate dative construction (generalization).}
  \label{fig:cdg}
  \vspace{-1em}
\end{wrapfigure}

\paragraph{Step 1: Define preconditions to test} We define three preconditions that we take as necessary to consider LM simulation studies as worthy of deriving hypotheses from. The first is exhibiting sufficient competence in English grammar, since the learning setup presupposes being able to comprehend the stimuli, which are simple English sentences of the kind appearing in child-directed speech. The second is exhibiting sensitivity to already-known English verb alternation patterns (e.g., that \textit{I donated them the book} is not acceptable but \textit{I delivered them the book} is acceptable), in order to verify that the models are capable of capturing the distinction we are interested in. The final is specifically being able to recognize that the novel word being taught in the simulated experiments has the category of verb---this also connects to the first precondition in that it evaluates the competence necessary to properly comprehend the stimuli sentences.

\paragraph{Step 2: Replicate known experimental results} Another important condition for the simulated learners to satisfy is that they replicate empirical findings of relevant prior work. We specifically aim to replicate the findings of lab studies of nonce word learning in children \citep{conwell2007early,arunachalam2017preschoolers} before exploring novel hypotheses, since these are the key word learning studies that we draw inspiration from to design the simulation experiments.

\paragraph{Step 3: Simulation of novel experiments} We treat language models trained on a corpus of age-ordered child-directed speech (AO-CHILDES \citep{huebner2021using}) as our simulated learners, and conduct model selection on the basis of Steps 1 and 2. The goal here is to conduct large-scale, systematic simulations of experiments that \textit{do not} exist in the literature. The simulation paradigm we use is visually illustrated in Figure~\ref{fig:cdg}, where we ``expose'' the language model to the new verb being learned (e.g., \textit{pilk}) via a single sentential context in one structure (DO or PO) and quantify generalization to the alternative structure via language model probability. We repeat this single-context learning experiments for all possible feature configurations of the exposure stimuli; because the only exposure each learner receives about the new verb is restricted to a single context, any difference in the model assigned probability to the generalization stimuli can only be attributed to the variation of the feature configurations of the context (e.g., the model learning \textit{pilk} from \textit{She pilked the ball to mommy} (nominal theme/recipient) vs. \textit{The red dog pilked me to the ball} (pronominal theme/nominal recipient)). We expand upon the technical details of this Step in three parts in \cref{sec:pipeline}.

\paragraph{Step 4: Hypothesis generation}
As discussed above, our hypothesis space is defined by a set of distributional features collected from the literature. Our approach for hypothesis generation is to run simulation studies that quantify the effect of all plausible combinations of these features on LM learners' generalization through statistical analyses, and then formulate hypotheses based on the results with the ultimate goal of testing them with human subjects \revision{in collaboration with language acquisition researchers}. 

\section{Related Work}
\label{sec:related-work}


\subsection{Neural Nets for Language Learning, Processing, and Generalization}

\revision{Our work joins the long tradition of using neural networks as computational tools to explore language learning, processing, and generalization. This tradition started in the 1990s due to Jeff Elman's foundation work in developing the recurrent neural network \citep[RNN;][]{elman1990finding}, alongside early psychological explorations of learning \citep{elman1993learning} and sentence comprehension \citep{mcclelland1989sentence, john1990learning}. More recent instances of such explorations have taken multiple different forms.}

\revision{A particularly productive line of work uses surprisal measures and/or intermediate vector representations, extracted from neural network LMs trained on large corpora, to model behavioral signatures extracted from human psychometric \citep[][\textit{i.a.}]{goodkind-bicknell-2018-predictive, wilcox2020predictive, van2021single, oh2023does, shain2024large, huang2024large} and neurological measurements \citep[][\textit{i.a.}]{schrimpf2021neural, michaelov2022so, goldstein2022shared, li2023heuristic, michaelov2024strong}. Another flourishing area of research has sought to evaluate neural network LMs for a variety of linguistic phenomena using carefully constructed minimal pair sentences \citep{linzen-etal-2016-assessing, marvin-linzen-2018-targeted, warstadt2020blimp, gauthier-etal-2020-syntaxgym, hu-etal-2020-systematic}, starting with English, and extending to a number of different languages \citep{xiang-etal-2021-climp, someya-oseki-2023-jblimp, suijkerbuijk2025blimp, jumelet-etal-2026-multiblimp}. With the development and proliferation of a variety of interpretability methods, there has been a great deal of work in investigating how LMs represent linguistic structure internally \citep{ettinger-etal-2016-probing, hewitt2019structural, arora-etal-2024-causalgym, boguraev-etal-2025-causal, boguraev2026drawbridges}. Finally, an especially up and coming strand of research in this area has been to develop models that can learn from a cognitively plausible \textit{amount} of data, with the development of the BabyLM challenge \citep{warstadt-etal-2023-findings}. This has in turn also spurred a concurrent line of research called \textit{controlled rearing} \citep{misra-mahowald-2023-rare} or \textit{filtered corpus training} \citep{patil2024filtered}, where language models are trained on tractable amounts of data that have been systematically manipulated to test the effect of removing or modifying a specific set of constructions \citep[see also][]{kallini-etal-2024-mission}. These controlled training methods are closest to our work, in that they seek to establish which cues in a model's training data allow it to generalize to a target phenomenon.}

\revision{Nonetheless, due to the apparent success of LMs in many of the aforementioned investigations, there has been much excitement about the potential of linguistics or cognitive science to gain back or derive insights from simulations on neural networks \citep{baroni2020linguistic, linzen2021syntactic, warstadt2022artificial, frank2023bridging, portelance-jasbi-2024-roles, futrell2025linguistics, frank2026cognitive, misra2026semantic}. To the best of our knowledge, these insights have not necessarily materialized yet, since a good chunk of the aforementioned literature has sought to corroborate results that already exist in humans. While performing such analyses is an important step (and is in fact a critical part in our framework, as mentioned in \cref{subsec:intro-case-study}), it is also important to generate \textit{novel}, empirically grounded predictions from these models, insofar as they \textit{do} satisfy the preconditions and prerequisites of the domain of interest. That is, we ought to build stronger bridges between studies on LMs and those on humans.
This uniquely positions our work in this literature, since we contribute a first step, in the form of a systematic framework for generating new, falsifiable hypotheses from simulations on LM learners, that are testable in the lab with human subjects.}

\subsection{Cognitive Modeling of Argument Structure Alternation} 

\revision{The two dative constructions we study fall under the broader umbrella of argument structure alternations \citep{goldberg1995constructions}, which have been a popular target for both Bayesian and Connectionist approaches \citep{alishahi2010computational}.}

\revision{For instance, \citet{alishahi2008computational} use an incremental Bayesian model to learn constructions by clustering individual scene-utterance pairs. Each pair specifies properties such as the semantic primitives of the event, the roles and semantic categories of its participants, and the ordering of the arguments. Their training data is synthetically generated using verb and argument structure frequencies estimated from the CHILDES corpus, and the model assumes that the scene and utterance have already been mapped onto these representations. It accounts for several known patterns in construction learning, but does not study generalization across dative constructions.}

\revision{\citet{perfors2010variability} also take a Bayesian approach, focusing specifically on the dative alternation. Their learner models the extent to which a novel verb generalizes to an unobserved construction, based on how it and previously encountered verbs are distributed across PO and DO---similar to the condition we test in this work. However, the source of evidence to this learner is restricted to count-based distributions of verbs across the different constructions and the alternation status of the verbs themselves (alternating, PO only, DO only). That is, their learner does not learn from more finer-grain distributional features of the individual exposures that the novel (and the previously encountered) verbs occur in.}

\revision{\citet{ambridge2016connectionist} take a connectionist approach to the dative alternation. Their model predicts whether a verb occurs in PO, DO, or another construction based on the verbs' lexical semantic properties, verb-construction frequencies, and the semantic representation of events (whether or not it involves \textit{transfer}; see \citet{hovav2008english}). It accounts for several effects observed in human judgments, but is trained on corpus-derived verb-construction frequencies and human ratings of verb semantics rather than directly on the sentences from which these properties would have to be learned.}

\revision{In each of the above cases, the learner’s input has already been analyzed into representations chosen for the phenomenon: whether these are construction counts, structured scene-utterance pairs, or lexical semantic features of the verbs. Our LMs are not constructed to \textit{explicitly} store or represent the construction category or the precise lexical semantics of individual items, and are trained on all kinds of utterances in the data, not just the phenomenon-relevant ones. Insofar as the aforementioned phenomenon specific information is being utilized by LMs, it can only arise as an emergent process of their training \citep{mcclelland2010letting} on input sequences as a result of their distributional objective of next-token prediction. Overall, our setup allows us to vary finer-grained properties of a novel verb's exposure while keeping its frequency and exposure construction fixed, which is not necessarily possible with the earlier approaches. This does not necessarily make LMs better models of human cognition---the earlier models make much clearer theoretical commitments about the learner's representations and learning mechanisms, while the interpretability of LMs remains an open question of active interest in computational cognitive science \citep{futrell2025linguistics}. We instead use LMs as simulated learners for identifying hypotheses that can be tested with humans, and view our contributions as complementary to these earlier approaches.}

\subsection{Dative Alternation and Language Models}
\revision{
Dative alternation has been frequently adopted as a testing ground for analyzing structural representations in LMs.
For instance, LMs such as GPT-2 \citep{radford2019language} and Llama-2 \citep{touvron2023llama} have been shown to exhibit effects analogous to structural priming \citep{bock1986syntactic} with respect to different dative structures \citep{sinclair-etal-2022-structural, jumelet2024language, zhou2024context}.
In these studies, LMs assigned greater probabilities to sentences in a particular dative construction when adapted to instances of the same construction---either via backpropagation or prefixes presented to the LM in context. 
Furthermore, \citet{hawkins-etal-2020-investigating} showed that BERT \citep{devlin-etal-2019-bert} and GPT-2 are sensitive to alternation patterns of known dative verbs \citep{levin1993english}, matching human alternation preferences. Analogous to humans, LMs also showed sensitivity to argument features such as length and definiteness which are known to predict speakers' dative choice during production \citep{bresnan2007predicting, de2012statistical}. This was further shown by \citet{yao2025both} who used the \textit{controlled rearing} method \citep{frank2023bridging, misra-mahowald-2023-rare} to show how sensitivities to length might originate from the general property of the target language (there, English) being ``short-first'' \citep{macdonald2013language}. \citet{ranganathan-etal-2025-semantic} showed masked language models such as BERT and RoBERTa \citep{liu2019roberta} to show sensitivity to the animacy of the theme and recipient arguments of dative construction. Specifically, they showed that the representation of \textit{London} as the recipient in the DO (\textit{I sent London the letter}) correlated better with animate properties (like `biomotion'), while that of the same word as the recipient in the PO form (\textit{She gave the letter to London}) correlated better with inanimate (especially, location) properties (like `landmark' and `scene'). Finally, \citet{rakshit-goldberg-2025-meaning} investigate internal representations of the Pythia-1.4B LM \citep{biderman2023pythia}, finding PO and DO sentences to be more distinct when human judgments strongly favor one construction over the other. This suggests that the model's representations reflect graded preferences tied to the verbs and arguments in those sentences.}

\revision{Some studies have also investigated dative alternation in LMs from the perspective of novel word learning \citep{thrush2020investigating, petty2022language}---these are the most directly related to our work in terms of methodology.
Specifically, \citet{thrush2020investigating} simulated novel verb learning in BERT, and showed that the model can use novel verbs in alternate constructions, having only been exposed to one of the constructions. This work masked out all content words during the course of BERT's finetuning, opting to instead only expose skeletons of alternating constructions (e.g., to expose BERT to the DO, they used an input like: \textit{The [MASK] \textbf{daxed} the [MASK] a [MASK]}). As a result, the precise role of the features of the theme and recipient on the model's productive generalization of the novel verb (here, \textbf{\textit{daxed}}) is unclear. \citet{petty2022language} adopted a similar novel word learning method, but the word being learned was not a verb---instead, it was either the theme or the recipient in a dative sentence. Among other findings, they showed that LMs such as BERT and RoBERTa \citep{liu2019roberta} can learn to infer a novel word's argumenthood (theme vs. recipient) based on the position it occurs in a dative exposure, and that recipients are more likely to be animate than inanimate, echoing prior corpus analyses \citep{bresnan2007predicting}.}

\revision{While all these findings paint a convincing picture about LMs' learning of fine-grained properties of the dative alternation, they leave open whether the properties of a particular exposure affect how a model generalizes a novel verb---i.e., they do not test for cross-dative generalization. More importantly, all of them focus on effects already known to exist in humans or properties already known of dative alternation. Apart from testing LMs for basic knowledge of dative alternation, our work contributes results that enable the generation of new, testable hypotheses.
}

\section{Methods}

\subsection{Learner Modeling}

\subsubsection{Data}
We used AO-CHILDES \citep{huebner2021using} as the training dataset for our LM learners.
This corpus contains approximately 5M words from the American English portion of CHILDES \citep{macwhinney2000childes}, including children from 0 to 6 years of age.
It has been filtered to include only the child-directed portion, and is ordered temporally---details concerning the corpus pre-processing and filtering can be found in \citet{huebner2021using}.
Assuming that a soft upper-bound for the amount of linguistic input to an English speaking American child is around 1M words per month \citep{hart2003early, roy2015predicting,  dupoux2018cognitive, frank2023bridging}, the AO-CHILDES corpus consists of about 14\% of the linguistic input to a 3 year old.
We used the train/validation/test splits of AO-CHILDES provided by the BabyLM challenge \citep{warstadt-etal-2023-findings}, a competition for building LMs trained on developmentally plausible amounts of data.
There are 4.21M words in the training set (11.7\% of the soft upper bound of the linguistic input to an English speaking American child), 400K words in the validation set, and 340K words in the test set. All utterances in AO-CHILDES are in lowercase.

\subsubsection{LM architecture and training}
The LMs used in this work are autoregressive, decoder-only Transformers \citep{radford2019language} trained using the next-word prediction objective---specifically, the OPT architecture \citep{zhang2022opt}. However, the main novel verb learning method is agnostic to the model architecture.
We experimented with multiple hyperparameters in order to arrive at our final model. Specifically, we varied optimizer learning rate (0.001, 0.003, 0.0001, 0.0003), vocabulary size (8192, 16384),  number of layers (8, 16), number of attention heads (8, 16), hidden state dimension (256, 512), and feed-forward hidden dimension (1024, 2048). Our final model (selected using the criteria described in our model selection experiments, \Cref{sec:model-selection}) used 8 attention heads, 8 layers, a vocabulary size of 8192, an embedding size of 256, and a feedforward hidden dimension of 1024, with a learning rate of 3e-3.
This amounts to a total of 8.4M trainable parameters.
For the tokenizer, we followed BabyBERTa \citep{huebner-etal-2021-babyberta}---an LM trained on AO-CHILDES that showed strong performance on tasks targeting the linguistic competence of English-learning children---and used a Byte-Pair Encoding-based tokenizer \citep{sennrich-etal-2016-edinburgh}. We re-trained this tokenizer on our training set since the original BabyBERTa tokenizer was trained on a combination of corpora not limited to AO-CHILDES \citep[see][]{huebner-etal-2021-babyberta}, leading to superfluous tokens not in our training corpus and therefore irrelevant to our investigation. We trained our LMs on the training set of the AO-CHILDES dataset for 10 epochs, and chose the best learning rate based on the validation set.
To ensure that our results are not due to idiosyncrasies of a particular training run, we report results for 5 different random seeds (i.e., 5 instances of our LM architecture trained on the same corpus using the same tokenizer, only differing by the initialization of the weights). 
Details about the implementation and open model release can be found in Section 1 of the Supplementary Materials.

\subsection{Simulation Pipeline}
\label{sec:pipeline}
\begin{figure}[!t]
    \centering
    \begin{subfigure}[t]{\textwidth}
    \centering
        \caption{\label{fig:exp-design}}
        \includegraphics[height=6cm]{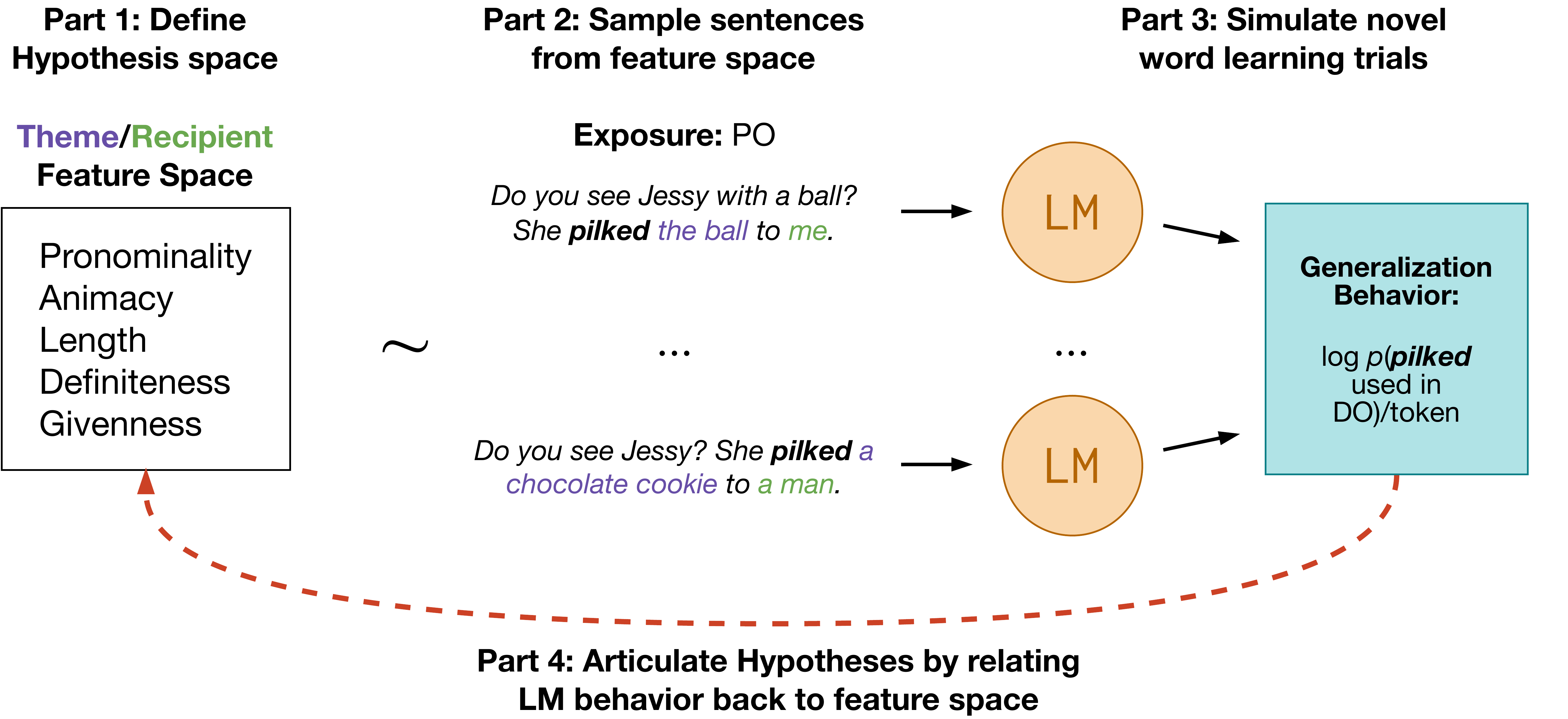}
    \end{subfigure}
    \vspace{1em}
    \begin{subfigure}[b]{\textwidth}
    \centering
        \caption{\label{fig:learning-trial}}
        \includegraphics[width=\textwidth]{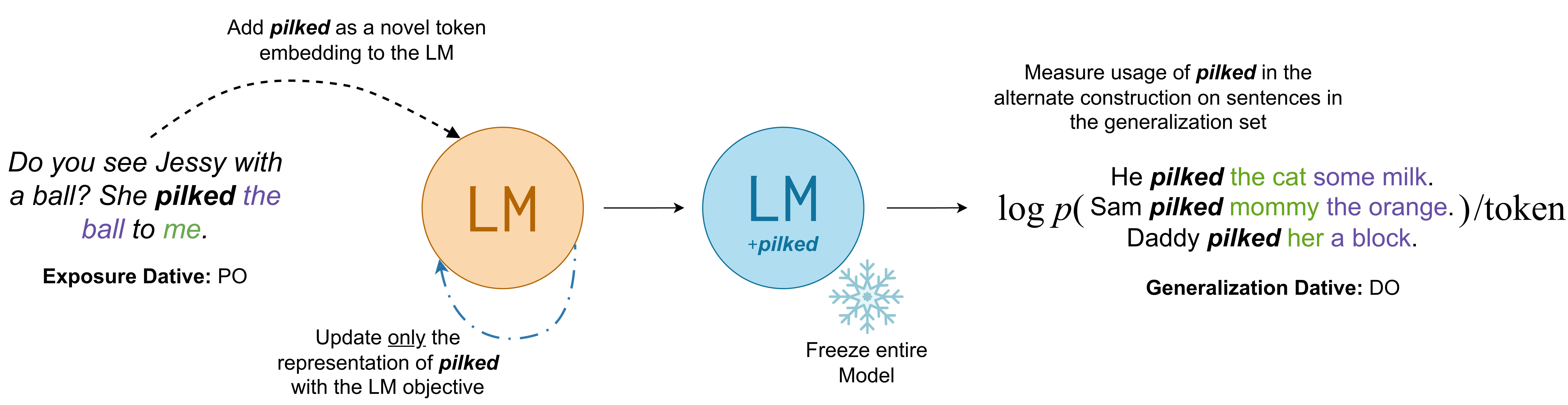}
    \end{subfigure}
    \caption{
    Overview of our experimental setup.
    \textbf{\textsf{a.}} Our simulation pipeline. We accumulate the LM learners' cross-dative generalization behavior, given exposure to a range of stimuli sampled from a space of theme and recipient feature configurations. We then relate this behavior back to the feature configurations themselves to articulate \textit{novel} hypotheses about the properties of the training exposure that facilitate cross-dative generalization. \textbf{\textsf{b.}} Depiction of a single learning trial, where a model's representation of a novel verb, \textit{\textbf{pilked}}, is updated given exposure to its use in a single dative sentence, following which the model is then frozen and evaluated on its behavior on a set of stimuli where \textit{\textbf{pilked}} is used in the unmodeled dative construction.}
    \label{fig:experiment-setup}
\end{figure}

This section describes our simulation pipeline setup, following the structure laid out in Figure~\ref{fig:experiment-setup}.

\subsubsection{Part 1: Define the hypothesis space (Theme/Recipient feature combination)}
\label{sec:hypothesis-space}

Our choice of the features of the exposure contexts is motivated by prior work characterizing dative alternation in human learners via production and comprehension experiments \citep{conwell2007early, rowland2010role, rowland2014competition, stephens2015dative, arunachalam2017preschoolers, conwell2019effects}, as well as corpus analyses of child-directed speech and adult-adult conversations \citep{gropen1989learnability, bresnan2007predicting, de2012statistical}. 
These studies have primarily focused on the features of the theme and recipient, since a core difference between the two constructions is the ordering of these arguments. 
Inspired by these previous works analyzing the usage of datives, we explore a large space of the possible feature configurations of the theme and recipient by constructing sentences containing a novel dative verb (the learning target). These sentences serve as exposure contexts in our simulation trials that the LMs learn the novel verb from.
We consider the following set of features for each argument: (1) \textbf{Pronominality}: whether the argument is a pronoun (e.g., \textit{her, it}, etc.) or a full noun phrase (e.g., \textit{the cat, a glass of milk}, etc.); (2) \textbf{Animacy}: whether the argument is animate (e.g., \textit{mommy, a boy}, etc.) or inanimate (e.g., \textit{something, the ball}, etc.); (3) \textbf{Definiteness}: whether the argument is definite (e.g., \textit{the man, me}, etc.) or indefinite (e.g., \textit{a girl, someone}); (4) \textbf{Givenness}: whether the argument has been established in previous discourse (given) or not (new); (5) \textbf{Length}: the number of words in the argument (e.g., ``\textit{the chocolate cookie}'' has three words), \revision{following prior corpus analysis and language model work  on the dative alternation \citep{bresnan2007predicting, de2012statistical, hawkins-etal-2020-investigating, yao2025both}.} A subset of these features (in particular, length, definiteness, and animacy) have been shown to be captured by off-the-shelf pre-trained LMs \citep{hawkins-etal-2020-investigating, ranganathan-etal-2025-semantic} as well as those trained on a developmentally plausible amount of data \citep{yao2025both}. More importantly, these features have shown to correlate with alternation choices in adult and child-directed speech corpora \citep{bresnan2007predicting, de2012statistical}, indicating that they are available to the learners as distributional cues (although, again note that the mere existence of distributional cues do not entail that the cues are actually used by the learners).

\subsubsection{Part 2: Sample exposure stimuli from feature space}

Our exposure stimuli for the LM learners are sampled from the feature space defined above.
Each item in our stimuli is a two-sentence utterance,  consisting of an agent, a novel verb (here, \textit{pilked}), the theme, and the recipient. The first sentence specifies what is given in the discourse---we consider an argument to be given if it is mentioned in the first sentence, and new otherwise. The agent (one performing the action of \textit{pilking}) is always mentioned here, whether or not the theme and the recipient are also mentioned is specified by the givenness feature configuration of the item. The second sentence specifies the dative construction (either DO or PO), and contains the agent, the novel past-tense verb \textit{pilked}, the theme, and the recipient. \Cref{ex:dative-templates} shows our stimuli templates.

\ex. \label{ex:dative-templates}
\a. \texttt{\{givenness-template\}}. \texttt{\{agent\}} \textbf{pilked} \texttt{\{recipient\}} \texttt{\{theme\}}. \textsc{[do]}
\b. \texttt{\{givenness-template\}}. \texttt{\{agent\}} \textbf{pilked} \texttt{\{theme\}} to \texttt{\{recipient\}}. \textsc{[po]}

A constraint common to all stimuli in our experiments is that all lexical items that will go in the above slots appear in our LMs' training set, AO-CHILDES.
For our agent slot, we use proper names.
We fill in the theme and recipients slots by exhaustively varying all features in our hypothesis space, and sampling lexical items that meet the criteria of each feature configuration. Specifically, we treat pronominality, animacy, definiteness, and givenness as binary features, and length as the difference between the length of the theme and that of the recipient, measured in number of words. Each binary feature is typically associated with its own set of lexical items (often overlapping across features---e.g., indefinite pronouns are restricted to \textit{it, something, someone}). We use adjectival (e.g., \textit{the red ball}) and prepositional (e.g., \textit{a boy with blue pants}) modification as a means to increase the diversity of our lexical items' lengths. In total, the difference between the theme and recipient lengths varies between -6 and 6 (i.e., 13 different values), which combined with 8 binary features (4 per argument) gives us 3328 possible feature configurations ($2^8 \times 13$). However, a majority of these configurations result in empty sets of lexical items---e.g., only a length difference of 0 is possible when both themes and recipients are pronominal, it is pragmatically infelicitous to have an item be given in the discourse but referred to using an indefinite article (e.g., \textit{Do you see Jenny with \textbf{the ball}? She pilked \textbf{a ball} to me.}), etc. On discarding such cases, we end up with a total of 756 possible feature configurations per dative construction, from which we sample lexical items to generate our stimuli items. An example item in the PO construction is given by \cref{ex:jenny}:

\ex. \label{ex:jenny} Do you see Jenny with the ball? Jenny pilked [it]$_{\textrm{theme}}$ to [a boy with blue pants]$_{\textrm{recipient}}$.\\
\textbf{theme:} \textit{pronominal, inanimate, given, definite}, \textbf{recipient:} \textit{nominal, animate, new, indefinite},\\ \textbf{length-difference:} $-4$

We sample 8 items for each feature configuration, using a different proper name for each sampled item as the agent. We experiment with three different surface form variations in the givenness template (i.e., the introduction sentence). This gives us 6048 items per givenness template per dative construction, yielding 36,288 items in total. Full details of stimuli generation can be found in Section 2.1 of the Supplementary Materials.

\subsubsection{Part 3: Simulate novel word learning trials with LM learners}

Our aim is to characterize the generalization behavior of an LM learner, given exposure to a novel verb in a particular dative construction. This analysis is conducted over a series of learning trials where in each trial the model will be exposed to the novel verb via a single sentence (with some preceding discourse context) in one dative construction, with a given feature configuration, and then tested for its usage of the novel verb in a collection of sentences in the alternate construction (Figure~\ref{fig:cdg}). 
As an example, the model can be exposed to the verb \textit{pilked} in \cref{ex:jenny}, and tested for generalization using sentences in \Cref{ex:genset}.

\ex. \label{ex:genset}
\a. You \textit{pilked} papa an apple, didn't you?
\b. I \textit{pilked} you a present.

To expose the model to a novel verb, we first add new token (here, \textit{pilked}) to the model's embedding layer as a randomly initialized vector. That is, the character sequence `pilked' is set to be a single token in the model's tokenizer, which then maps it to this newly added vector in the model's embedding layer. After adding the new token and its embedding, we pass the exposure stimulus (containing \textit{pilked}) to the model as input, and update the model's weights for a predefined number of steps using the same objective (i.e., next token prediction) and learning rate as the model's original training. While in principle the entire model could be updated, we choose to only update the newly added embedding vector and keep the rest of the pretrained weights of the model frozen, so as to preserve their effect on the novel token as it is being updated. 

Finally, to quantify the model's generalization behavior, we create two generalization sets, one for each dative construction. For this, we used natural dative sentences occurring in the validation and test sets of the AO-CHILDES corpus, with the verb being in the past tense, and replaced the verb with \textit{pilked}. We detected these sentences using a semi-automatic pipeline described in Section 3 of the Supplementary Materials, and found 160 sentences in the DO construction and 95 in the PO construction. To measure the generalization behavior of a model exposed to the novel verb in a given dative construction, we compute its average log probability per token assigned to sentences in the generalization set of the alternate construction. That is, if the model is exposed to \textit{pilked} in DO, then we compute the model's average log probability per token on the 95 sentences in the generalization set where \textit{pilked} was used in a PO construction.

The design decisions of our method are motivated from paradigms in developmental studies. Our method itself is adapted from \citet{kim2021testing}, who developed it to study lexical category inference in LMs, and were inspired by developmental studies studying an analogous question in children \citep{hohle2004functional} using a head-turn preference procedure \citep{nelson1995head}. Our experimental design of single sentence exposures follows from development work in novel verb acquisition \citep{conwell2007early, arunachalam2013out, arunachalam2017preschoolers}. Finally, our decision to measure generalization of the model's usage of the novel verb to the alternate form follows from previous work on children's production/comprehension of the verb in the \textit{unmodeled} form \citep{conwell2007early, arunachalam2017preschoolers}. \Cref{fig:learning-trial} depicts an example of a single learning trial.

\subsubsection{Part 4: Relate generalization outcomes back to hypothesis space}
Conducting learning trials on our entire set of stimuli yields a large collection of generalization behavior estimates, each associated with a particular input exposure, which in turn corresponds to a specific feature configuration. This then allows us to measure the effect on generalization behavior of a learner as a function of the changes to each feature in the input exposure, using a linear mixed-effects model. For example, we can now measure the effect of having a pronominal vs. a non-pronominal theme in a PO exposure on a learner's generalization to the DO construction. 
Here, input exposures that promote the usage of the novel verb in the unmodeled construction (i.e., show greater cross-dative generalization) will be associated with greater log probabilities per token on the generalization set, while those that are preemptive will be associated with lower values.
We will use this collection of model results as our primary evidence from which we will derive our hypotheses about the conditions (e.g., feature configurations) that enable cross-dative generalization.

\section{Results}

\subsection{Model Selection}
\label{sec:model-selection}
As discussed in Section~\ref{subsec:intro-case-study}, rigorous model selection is an integral part of our hypothesis generation pipeline (Figure~\ref{fig:pipeline}) because it strengthens our confidence that novel hypotheses derived from the models would be borne out with human subjects.

We carry out model selection based on the three preconditions discussed in \Cref{subsec:intro-case-study}: (1) grasp of basic English grammar; (2) sensitivity to known alternation patterns of existing verbs that only occur in a single dative construction in the models' training data; and (3) recognition of the novel word being learned as a verb.  We use preconditions (1) and (2) to tune LM hyperparameters, selecting the ones that best satisfy them. We use precondition (3) to select the best representational state of the novel verb being learned (see Section 4 of Supplementary Materials for details). Results from analyses focusing on preconditions 1 and 2 are discussed below:

\subsubsection{Targeted syntactic evaluation}
A popular method of testing LMs on their ability to capture syntactic phenomena is by evaluating them on minimal pair benchmarks \citep{marvin-linzen-2018-targeted, warstadt2020blimp}. These benchmarks consist of a collection of pairs of sentences, where one of them is grammatical and the other involves a minimal change to the original sentence which renders the sentence ungrammatical in a systematic, phenomenon-specific way. An example of one such minimal pair, which falls under the `subject-verb agreement' phenomenon is shown below:

\ex. \label{ex:mp}
\a. \label{ex:mp-correct} The cat \underline{\textit{is}} screaming for food.
\b. \label{ex:mp-incorrect} The cat \underline{\textit{are}} screaming for food.

Here, \cref{ex:mp-correct} is converted into an ungrammatical sentence \cref{ex:mp-incorrect} by replacing \textit{is} with \textit{are}, which disagrees with the number of the subject, \textit{cat}. An LM that has learned subject-verb agreement should assign greater probability to acceptable sentences like \cref{ex:mp-correct} than unacceptable ones like \cref{ex:mp-incorrect}. Accuracy is computed as the percentage of time an LM's log probability per token of the acceptable sentence was greater than that of the unacceptable sentence. 

\begin{figure}[!t]
    \centering
    \begin{subfigure}[t]{0.20\textwidth}
    \centering
        \caption{\label{fig:zorro}}
        \includegraphics[height=4.7cm]{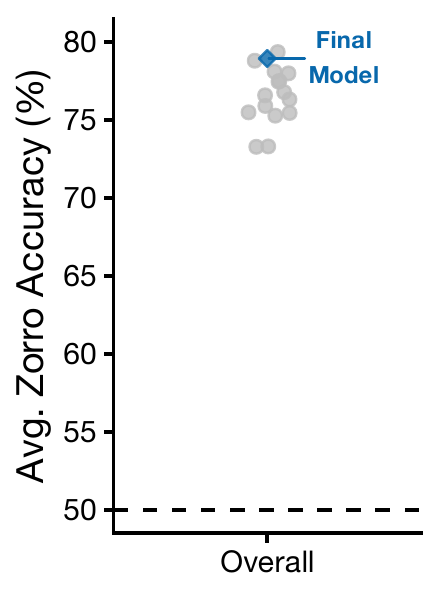}
    \end{subfigure}
    \hfill
    \begin{subfigure}[t]{0.20\textwidth}
    \centering
        \caption{\label{fig:nabanana}}
        \includegraphics[height=4.67cm]{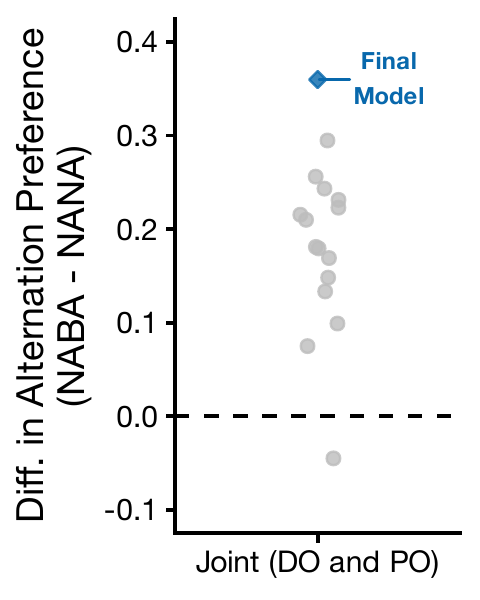}
    \end{subfigure}
    \hfill
    \begin{subfigure}[t]{0.55\textwidth}
    \centering
        \caption{\label{fig:nabanana-final}}
        \includegraphics[height=4.75cm]{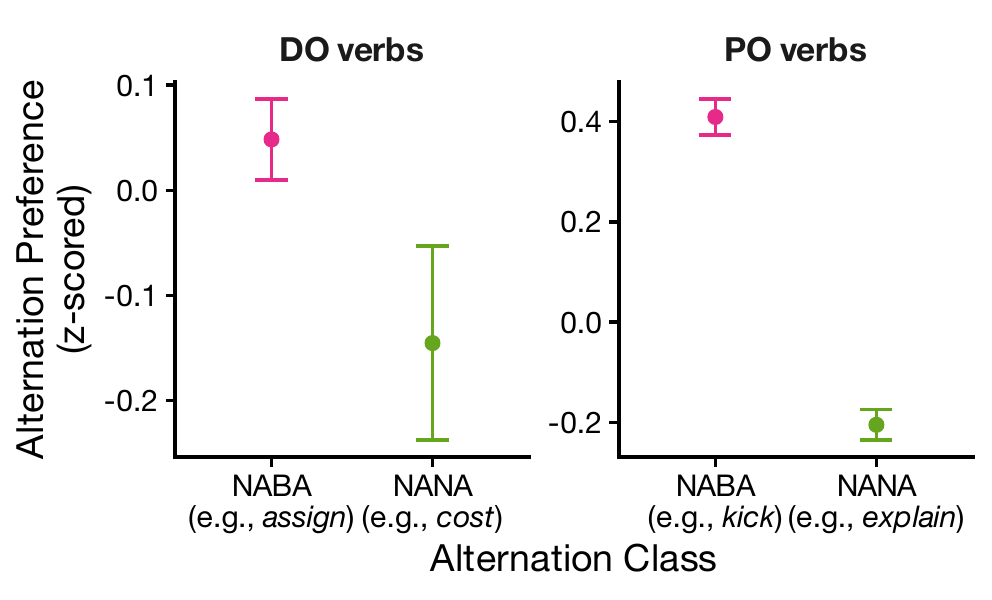} 
    \end{subfigure}
    
    \caption{Model selection results from evaluating the preconditions of grammar learning and sensitivity to known dative alternation patterns. \textsf{\textbf{a.}} Average accuracy (\%) of models with varying hyperparameters on Zorro \citep{huebner-etal-2021-babyberta}, a minimal pair dataset targeting 24 different morphosyntactic phenomena used to evaluate grammar knowledge of LMs trained on child-directed speech. Chance performance is 50\%. \textsf{\textbf{b.}} Results on comparing alternation preferences of the model on verbs that do not alternate in models' training data but are known to alternate in reality (\naba{}) vs. verbs that do not alternate in models' training data and also do not alternate in reality (\nana{}). 
    The plot shows the average difference in the alternation preference for \naba{} verbs (N=12) and \nana{} verbs (N=14) in AO-CHILDES \citep{huebner2021using} for both constructions (DO and PO), for models with varying hyperparameters. Larger difference indicates higher sensitivity to the alternation pattern difference. \textsf{\textbf{c.}} Alternation preference (z-scored) of 5 different runs of the final model architecture, which we use to perform our subsequent experiments on \naba{} and \nana{} verbs. Across both dative constructions, models preferred sentences in the alternate construction for \naba{} verbs relative to those for \nana{} verbs (DO: $\beta$ = 0.079, \textit{t}(2794) = 4.467, \textit{p} \textless{} .001; PO: $\beta$ = 0.291, \textit{t}(6294) = 24.150, \textit{p} \textless{.001}). This suggests that the final models show non-trivial sensitivity towards the dative alternation preferences of real, known verbs that occur in the training data.}
    \label{fig:prereq}
\end{figure}

In our experiments, we used the Zorro benchmark \citep{huebner-etal-2021-babyberta}, which is a collection of minimal pair stimuli spanning 24 different phenomena, and has been popularly used to evaluate models trained on child-directed speech \citep{yedetore-etal-2023-poor, mccoy2025modeling, padovani-etal-2025-child}. Importantly, it only uses lexical items that occur in AO-CHILDES, our training set, making it an appropriate testbed to evaluate our models compared to an alternative dataset such as BLIMP \citep{warstadt2020blimp}. \Cref{fig:zorro} shows the average accuracies across the 24 phenomena contained in Zorro. All LMs trained during our hyperparameter search achieve average accuracies between 73.3\% and 79.4\%, all of which are substantially above random chance (50\%). 

\subsubsection{Predicting alternation preferences of known verbs that do not alternate in training}

While the previous criterion targets LMs' ability to capture general syntactic phenomena, we now turn to a criterion that more closely targets the phenomenon of dative alternation. Here, we test the extent to which an LM predicts the alternation patterns of real verbs that have asymmetric distribution in the models' training data.
The learning scenario we target here is as follows. The learner observes two types of verbs which only occur in a single dative construction (say, PO). One of these two types of verbs---outside of the limited set of the learner's observations---is in fact far more permissive of alternation than the other. Then, in terms of the pure co-occurrence statistics of verb and construction, the two types of verbs are equivalent: they both only occurred in PO.
In such a learning scenario, does an LM show different patterns of behavior in its usage of these two types of verbs in the alternate form (here, DO), in a way that aligns with known alternation patterns?
Indeed, this is a scenario that is not entirely infrequent in AO-CHILDES, our training set. There are 26 different verb lemmas that only occur in one dative construction, which we found by using the semi-automatic detection pipeline described in Section 3 of the Supplementary Materials.
Out of these, 12 have been classified as ``alternating'' according to \citet{levin1993english} and \citet{hovav2008english}, while 14 have been considered ``non-alternating'', which we take to be indicative of the strength of the preference rather than hard categories, following \citet{bresnan2009gradience}.
If there are distributional cues present in the context that the verbs occur in (other than the DO or PO constructions themselves), and if the LMs can use these cues as learning signal, they should prefer the usage of the ``alternating'' verbs and disprefer the usage of the ``non-alternating'' in their alternate forms.
For example, consider the following two contexts taken from the AO-CHILDES training set, containing the verbs \textit{deliver} and \textit{said}, both of which only occur in the PO construction:

\ex. \label{ex:nabanana-obs}
\a. \label{ex:naba-obs} you \textbf{delivered} mail to me and to gabby and debbie ?
\b. \label{ex:nana-obs} you \textbf{said} goodbye to part of the train ?

A learner that has recognized the fact that \textit{deliver} is more likely to alternate than \textit{said} would accept \textit{delivered} in both \cref{ex:naba-unobs-po} and \cref{ex:naba-unobs-do}, while \textit{said} only in \cref{ex:nana-unobs-po}:

\ex. \label{ex:naba-unobs}
    \a. \label{ex:naba-unobs-po} she \textbf{delivered} the box to them . 
    \b. \label{ex:naba-unobs-do} she \textbf{delivered} them the box .

\ex. \label{ex:nana-unobs}
    \a. \label{ex:nana-unobs-po} they \textbf{said} hello to me .
    \b. \label{ex:nana-unobs-do} ?they \textbf{said} me hello .

To test whether this pattern holds for LMs, we compared the behavior of LMs for pairs of sentences such as \cref{ex:naba-unobs} to \cref{ex:nana-unobs}. For brevity, we denote verbs that are not observed to alternate but tend to alternate (according to \citeauthor{levin1993english}'s classification) as \nabas{} (\textbf{N}ot \textbf{A}lternating in the training data \textbf{B}ut actually \textbf{A}lternating)---e.g., sentences in \cref{ex:naba-unobs}, while those that do not alternate according to \citeauthor{levin1993english} as \nanas{} (\textbf{N}ot \textbf{A}lternating in the training data and actually \textbf{N}ot \textbf{A}lternating)---e.g., sentences in \cref{ex:nana-unobs}. 
If on average, LMs' preference for the alternate form is greater for \naba{} verbs than it is for \nana{} verbs, then this would suggest a non-trivial role of the distributional cues accompanying each type of verb in teasing apart their tendencies to be used in the alternate form.
In other words, the distributional cues that accompany \naba{} verbs could promote their usage in the alternate form.
Similarly, the cues with which \nana{} verbs occur might serve as signals that demote the usage of the verb in the alternate form.

To test LMs on their alternation preferences for \naba{} and \nana{} verbs, we manually create a collection of minimal pair sentences that used a given verb in either dative construction. In total, we created 840 pairs of sentences for \naba{} verbs, and 960 sentences for \nana{} verbs.  See Section 5 of the Supplementary Materials for details about the minimal pair dataset construction. For each pair, we computed the difference in log probability per token of the unobserved form, and that of the observed form. 
For an LM to have learned the alternation behavior of our verbs, this difference should be smaller on average for \nana{} verbs than for \naba{} verbs, since the unobserved form should be more unlikely for \nana{} verbs---i.e., the LM should \textbf{dis}prefer sentences like \textit{``He \textbf{described} my uncle the day''} over ones like \textit{``He \textbf{described} the day to my uncle''}.

\Cref{fig:nabanana} shows the difference in alternation preference of \naba{} and \nana{} verbs, averaged across DO and PO constructions, for LMs with varying hyperparameter configurations. We see more variability here than in our Zorro results, suggesting that capturing this phenomenon could be more non-trivial and nuanced than learning general syntactic knowledge. We chose the final model by selecting the LM that maximized the product of its zorro accuracy and the joint difference between alternation preferences for \naba{} and \nana{} verbs. This model achieved the second-best accuracy on Zorro (0.4 percentage points worse than the best one), and the best difference on the \naba{}-\nana{} test. 
\Cref{fig:nabanana-final} shows the breakdown of average alternation preference of five different runs of the final LM, across both dative constructions, and alternation classes. We find our LM learners prefer unmodeled alternations of \naba{} verbs than of \nana{} verbs across both constructions (DO: $\beta$ = 0.079, \textit{t}(2794) = 4.467, \textit{p} \textless{} .001; PO: $\beta$ = 0.291, \textit{t}(6294) = 24.150, \textit{p} \textless{.001}).

\subsection{Replication of known cross-dative generalization results}

Before conducting the full experiment that explore the large hypothesis space laid out in \cref{sec:hypothesis-space}, we first test if our LM learners reflect existing patterns in children's cross-dative generalization of novel verbs \citep{conwell2007early, rowland2010role, arunachalam2017preschoolers}. These experiments are critical because the ``animal models'' approach we take relies on expectations of high behavioral alignment between LM and human learners. Hence, being able to replicate known findings about child learners is a precondition that needs to be met before exploring novel hypotheses with the ultimate goal of testing human learners. 

\revision{We take successful replication to mean replicating the directionality of the significant effects reported in the child studies, which use production \citep{conwell2007early} or comprehension tasks \citep{rowland2010role, arunachalam2017preschoolers}. This does not mean that we take our main measure, contrasting probabilities, as analogous to either production or comprehension. In both human and model studies, the phenomenon targeted is cross-structural generalization. Comprehension and production measures are how the developmental studies measure generalization to unmodeled structure, and probability comparison is how our study measures it. These are two operationalizations for measuring the same target, and therefore does not require interpreting our probability comparisons as a measure of either production or comprehension. Still, it allows us to examine whether the model exhibits generalization patterns consistent with ones observed in children.}

For details of the statistical analyses of each experiment, we refer the reader to Section 6 of the Supplementary Materials.

\subsubsection{Asymmetric cross-dative generalization}

Our first experiment focuses on replicating the findings of \citet{conwell2007early}. Their study found evidence for abstract, productive knowledge of the dative alternation in children (as young as 3;0) when they were exposed to a novel verb in a single dative construction. However, their productive generalization was \textit{asymmetric}: children productively used a novel verb in a PO construction when it was exposed to them in a DO construction, but withheld generalization to DO when exposed to the novel verb in a PO construction. That is, children were substantially more likely to generalize from DO to PO than from PO to DO. We tested our LM learners under similar cross-dative exposure and generalization conditions. Specifically, we tested if the LMs found sentences involving the novel verb used in the alternate construction more likely when originally exposed to the novel verb in a DO construction than in the PO construction. We did this by comparing the average log probabilities per token assigned by the LM learners to generalization set sentences in the alternate construction: i.e., if the LM was exposed to the novel verb in the DO construction, we measured the average log probability on PO sentences in the generalization set, and vice versa. We conducted this experiment using our full set of stimuli described earlier.

\begin{figure}[!t]
    \centering
    
    \begin{subfigure}[b]{0.30\textwidth}
        \caption{\label{fig:conwell}}
        \includegraphics[width=\textwidth]{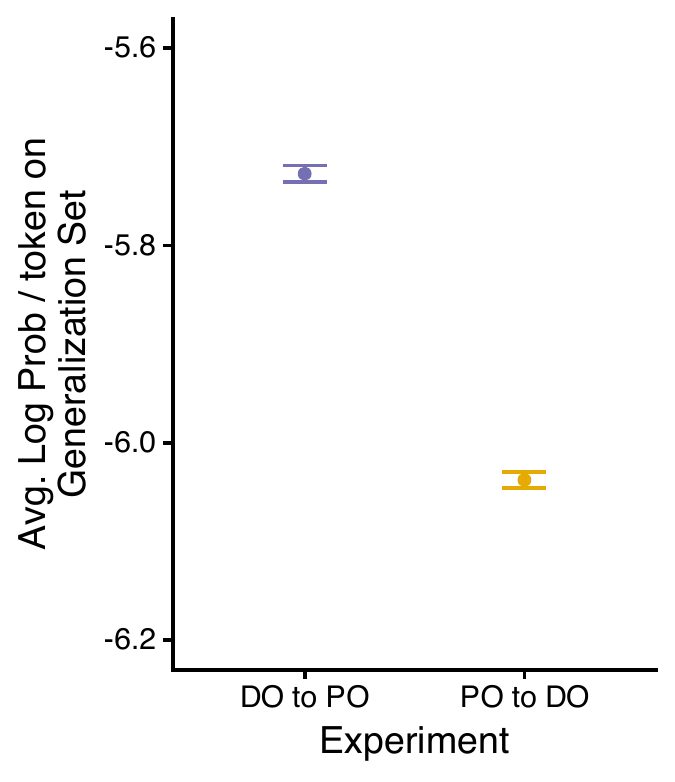} 
    \end{subfigure}
    \hfill
    \begin{subfigure}[b]{0.30\textwidth}
        \caption{\label{fig:arunachalam1}}
        \includegraphics[width=\textwidth]{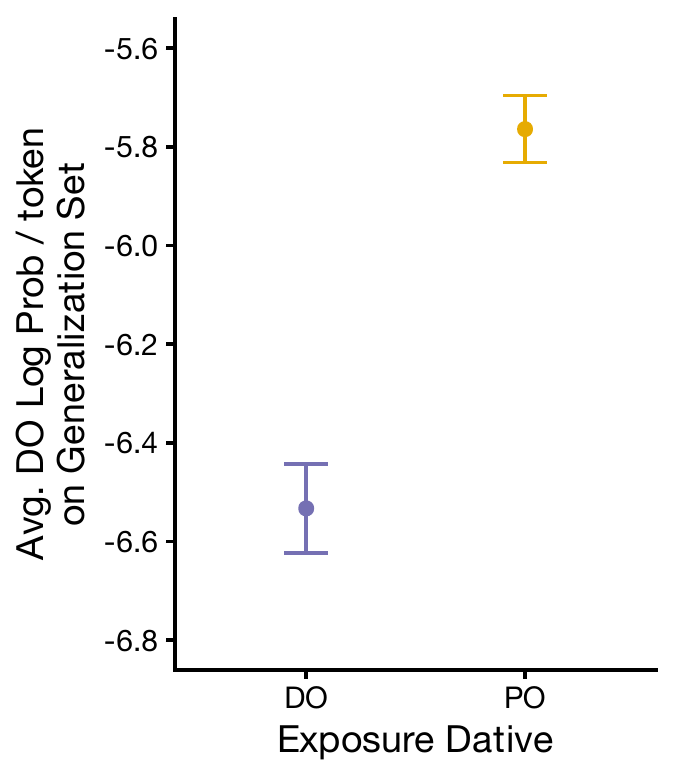} 
    \end{subfigure}
    \hfill
    \begin{subfigure}[b]{0.30\textwidth}
        \caption{\label{fig:arunachalam2}}
        \includegraphics[width=\textwidth]{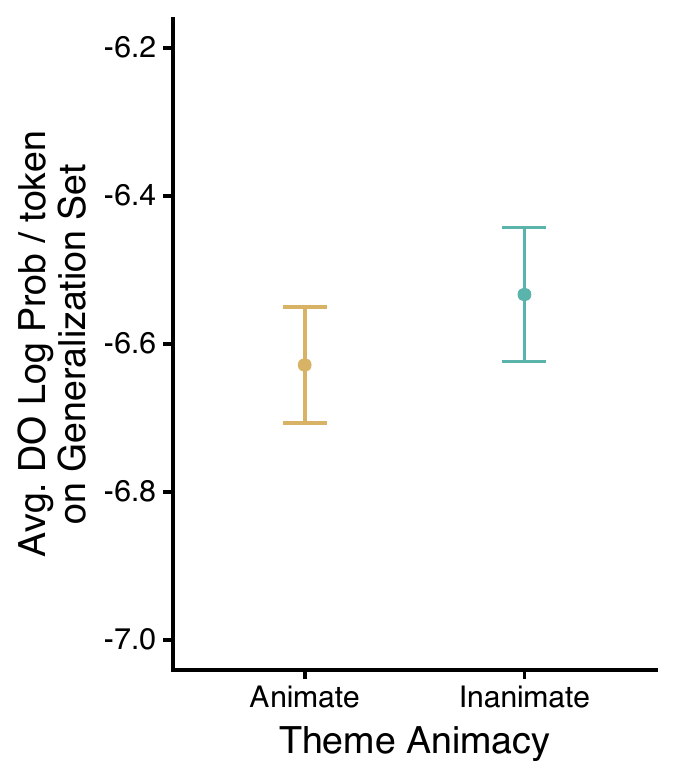} 
    \end{subfigure}
    
    \caption{Results from our replication of \textit{known} cross-dative generalization results \citep{conwell2007early, rowland2010role, arunachalam2017preschoolers}. \textbf{\textsf{a.}} Asymmetric cross-dative generalization \citep{conwell2007early}: Average log probabilities per token of LM learners on the generalization set was greater in the DO to PO experiment than in the PO to DO experiment ($\beta$ = 0.31; \textit{t}(4.15) = 6.11; \textit{p} \textless{} .01); \textbf{\textsf{b.}} Advantage of cross-structure training \citep{arunachalam2017preschoolers}: Average log probabilities per token of LM learners on the DO generalization set was greater when the exposure to the novel verb was a DO dative than when it was a PO dative ($\beta$ = 0.77; \textit{t}(3.99) = 7.39; \textit{p} \textless{} .01); \textbf{\textsf{c.}} Non-effect of theme animacy in learning from DO exposures \citep{rowland2010role, arunachalam2017preschoolers}: There was no significant effect of theme animacy in LM learners' average log probabilities per token on the DO generalization set, given exposure to the novel verb in a DO dative ($\beta$ = 0.09; \textit{t}(2.81) = 0.59; \textit{p} = 0.59).}
    \label{fig:main}
\end{figure}

\Cref{fig:conwell} shows the average log probabilities of the alternate constructions in the generalization set, as assigned by LM learners across different exposure conditions (DO vs. PO). LM learners exhibited relatively greater tendency to generalize from DO to PO than from PO to DO ($\beta$ = 0.31; \textit{t}(4.15) = 6.11; \textit{p} \textless{} .01). This asymmetric preference for DO to PO over PO to DO generalization aligns with children’s generalization pattern observed in the original study by \citet{conwell2007early}.

\subsubsection{Advantage of cross-structure training} 

Next, we test the role of different dative constructions (PO vs. DO) as exposures when learning a novel verb's usage in the DO construction. Prior work on testing this in children \citep{arunachalam2017preschoolers} has reported that the comprehension of a novel verb in the DO construction is more likely to be facilitated when it was exposed to the learner in a PO construction than when it was exposed to them in a DO construction with the same arguments. That is, there was an \textit{advantage of cross-structure training} \citep{rowland2010role, arunachalam2017preschoolers}, where exposure to novel verbs in easier-to-parse constructions (PO) facilitated subsequent processing in more syntactically difficult constructions (DO), echoing the results of prior work in novel word acquisition \citep{arunachalam2013out}. In the context of our LM learners, we tested whether similar observations were borne out by testing whether their log probabilities of using the novel verb in unseen DO sentences was greater when they were exposed to the novel verb in the PO than when they were exposed to it in the DO.

In the experimental stimuli of the original study \citep{arunachalam2017preschoolers}, the recipients and themes were both definite, discourse given, and short (two-word noun phrases). They differed only in their animacy, with themes being inanimate and recipients being animate. However, because this only corresponds to a single hypothesis configuration in our exposure stimuli, our sampling yields only a total of 16 test items. Therefore, for the purpose of this analysis, we included additional lexical items and expand the set of stimuli to total of 220 items per exposure construction (DO and PO). All items still adhere to the target feature configuration. Details of the stimuli expansion can be found in Section 2.2 of the Supplementary Materials.

\Cref{fig:arunachalam1} shows the average log probabilities per token of the LM learners on the DO generalization set (i.e., sentences with the verb used in an unseen DO context) across the two types of exposure contexts (DO vs. PO). We find a significant main effect of the exposure construction, with PO exposures resulting in greater generalization of the novel verb to the DO construction than did DO exposures ($\beta$ = 0.77; \textit{t}(3.99) = 7.39; \textit{p} \textless{} .01). That is, LM learners, like child learners, showed a cross-structure training advantage.

\subsubsection{No effect of theme animacy in learning from DO exposure}
Our final replication involves testing whether the animacy of the theme has an effect in learning novel verbs from DO exposures. In child learning experiments, \citet{arunachalam2017preschoolers} reported that for children (3;0--4;0), comprehension of a novel verb in the DO construction is equally likely for animate and inanimate themes (with animate recipients) when learning about the verb from DO exposures. Insofar as our LM learners capture this behavior, we expect there to be no significant effect of theme animacy in the DO subset of the generalization set log probabilities, given exposure to the novel verb in DO stimuli. 

We use the same dataset as the previous analysis on cross-structure training, but filter it to only include the DO to DO subset, and this time also include stimuli with animate themes in the experiment. Specifically, since the previous stimuli consists of 220 items in the DO to DO experiment with inanimate themes, we added another set of 220 items with animate themes, keeping all other items (e.g., the agent and recipient) the same. Then, we test if models' average log probabilities per token on the DO generalization set is significantly different when exposed to a novel verb in the DO construction with an animate theme vs. with an inanimate theme ($N$=220 exposures in each), keeping variation in all other aspects constant. 

\Cref{fig:arunachalam2} shows the average log probabilities per token of the LM learners on using the verb in the sentences of the DO generalization set, across all five random seeds. Analogous to the combined results of \citet{rowland2010role} and \citet{arunachalam2017preschoolers} on children, we found no significant effect of the animacy of the theme in the LM learners' generalization of a novel verb from exposure in a given DO construction to unseen DO instances ($\beta$ = 0.09; \textit{t}(2.81) = 0.59; \textit{p} = 0.59).

\subsection{Full Simulation}
\label{sec:full-simulation}

Having established that our model replicates key findings in prior studies, we now turn to our main simulation where we relate the cross-dative generalization behavior of LM learners to the feature configurations of the new dative verb's exposure. Results from this simulation serve as the primary evidence for articulating novel hypotheses about the exposure conditions that facilitate or preempt cross-dative generalization in human learners. 

Our experiments primarily varied the features of the theme and recipient of the exposure dative constructions. Given the large number of features, rather than fitting a statistical model with all features as predictors (as well as their interactions within and across arguments), we explore using a scoring scheme that aggregates the features and using the score as the main predictor. One natural way the features can be combined is based on the idea of \textit{Harmonic Alignment} in Optimality Theoretic Syntax \citep{prince1993optimality, aissen1999markedness, aissen2003differential}. In general, argument features have a tight relationship with the information structure of a sentence; related to our current discussion, English speakers' choice of dative construction in particular have been shown to be modulated by combinations of theme and recipient features \citep{thompson1990information, collins1995indirect, goldberg1995constructions, arnold2000heaviness, bresnan2007predicting, de2012statistical}. 
For example, speakers prefer DO over PO if the recipient is pronominal \citep{bresnan2007predicting}, speakers tend to place shorter and more discourse accessible entities (i.e., ones that have already been established in previous discourse, or given) before those that are longer and discourse-new \citep{arnold2000heaviness, arnold2008put, wasow2002postverbal}, etc. Harmonic alignment offers a explanation for these patterns, analyzing the preference as stemming from the degree of alignment between discourse prominence of the features (e.g., pronouns are more discourse-prominent than non-pronouns) and positional prominence (appearing earlier is more prominent than later). Under Harmonic Alignment, the combination of multiple prominent dimensions (and multiple less prominent dimensions) are considered \textit{harmonic}, and a combination of more prominent and less prominent dimensions are considered less \textit{harmonic}, and speakers prefer harmonic expressions to less harmonic expressions. In dative usage in child directed speech and adult-conversations, the scales \textit{given \textgreater{} new, pronominal \textgreater{} nominal, definite \textgreater{} indefinite, and short \textgreater{} long} have shown to align with the positional prominence scale for linear positions of the argument following the verb (i.e., \textit{earlier \textgreater{} later}) \citep{bresnan2007predicting, de2012statistical}. In designing a scoring scheme based on harmonic alignment, we adopt the aforementioned discourse prominence scales for all of our features, except for animacy. Regarding animacy, while there is a preference for animates to precede inanimates in more general production data \citep{aissen2003differential}, empirical work on dative constructions has \textit{only} observed this for adults, and further only restricted to the DO construction \citep{bresnan2007predicting}. This discrepancy between animacy and other features is likely affected by the fact that themes and recipients are prototypically inanimate and animate \citep{hovav2008english, beavers2011aspectual}, respectively (the intuition is that, animate things are more likely to receive things, and inanimate things are more likely to be given). Hence, we use animacy prototypicality in lieu of the more general harmonic alignment based on discourse prominence for the animacy feature. 

The scoring scheme to measure the extent to which an exposure stimulus conforms to the harmonic alignment, which we name \textbf{Harmonic Alignment and Animacy Prototypicality} (\haap{}) score, is as follows. The scoring scheme is designed so that stimuli that satisfy more alignment constraints are scored higher than those that satisfy fewer constraints. The \haap{} score is the sum of three components: one component each for the binary feature values of the two arguments (called theme and recipient scores), and one for the difference in arguments' lengths ($\Delta$length). For binary features of theme and recipient, we assign a score of 1 every time the feature is compliant with the expected features according to harmonic alignment and animacy restrictions, and 0 otherwise. This way, if the features of an exposure stimulus is perfectly in compliance, then it will receive a score of 4 for each argument, totaling to 8. We measure $\Delta$length using a sign-preserving log-transform \citep[following][]{bresnan2007predicting}.

\begin{figure}[!t]
    \centering
    \begin{subfigure}[t]{0.69\textwidth}
    \centering
        \caption{\label{fig:do-po-comparison}}
        \includegraphics[height=5cm]{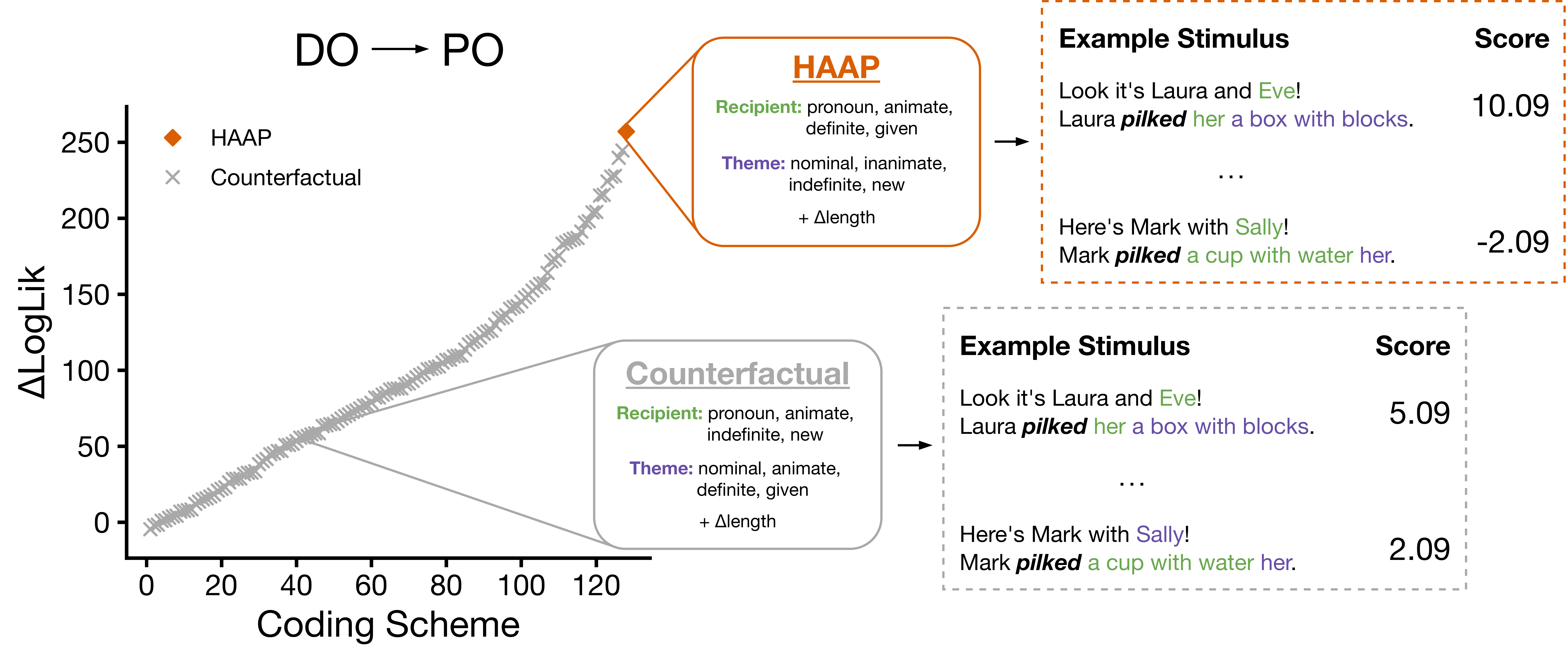}
    \end{subfigure}
    \hfill
    \begin{subfigure}[t]{0.28\textwidth}
    \centering
        \caption{\label{fig:do-po-lmer-haap}}
        \includegraphics[height=5cm]{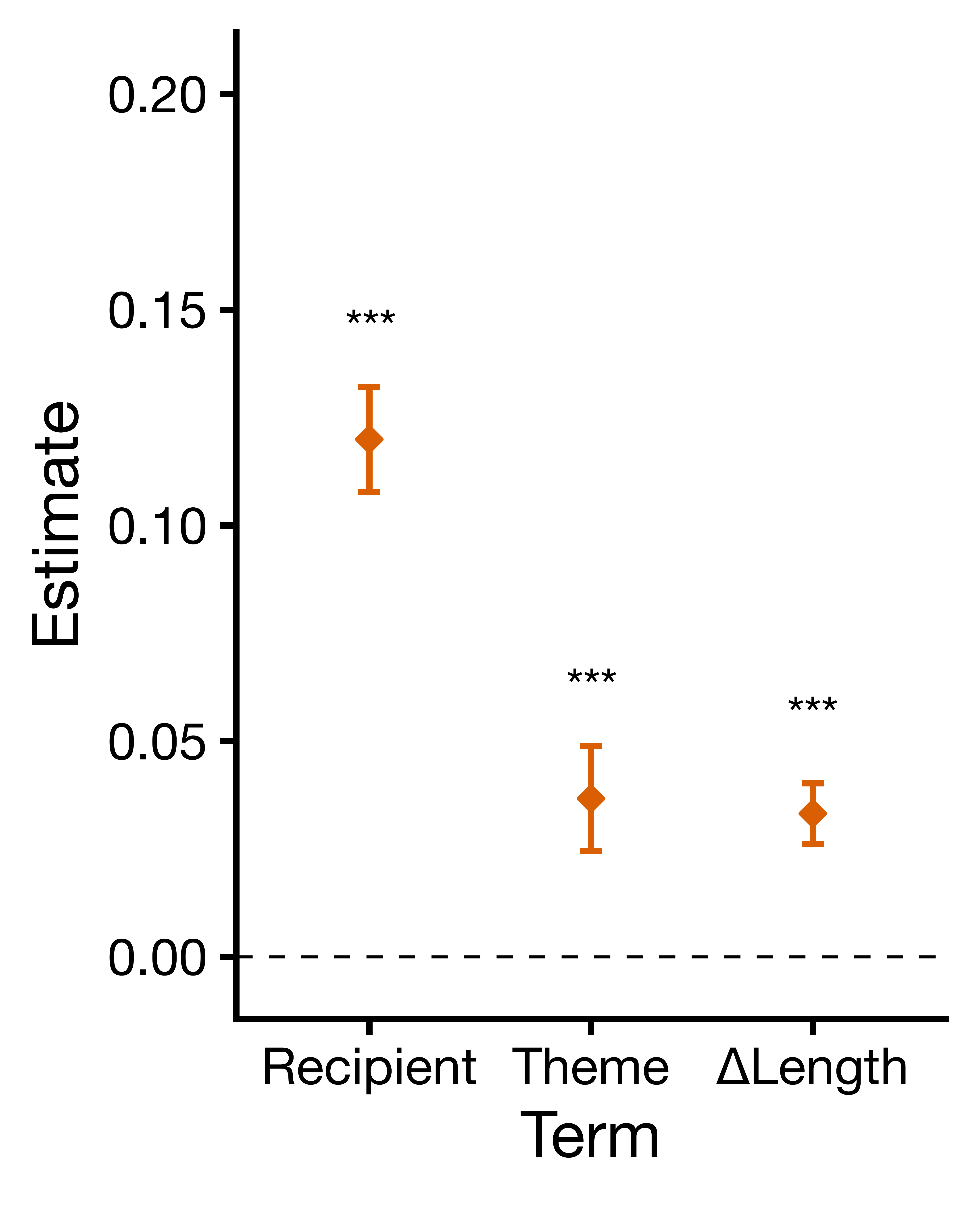}
    \end{subfigure}
    \vspace{1em}
    \begin{subfigure}[t]{0.69\textwidth}
    \centering
        \caption{\label{fig:po-do-comparison}}
        \includegraphics[height=5cm]{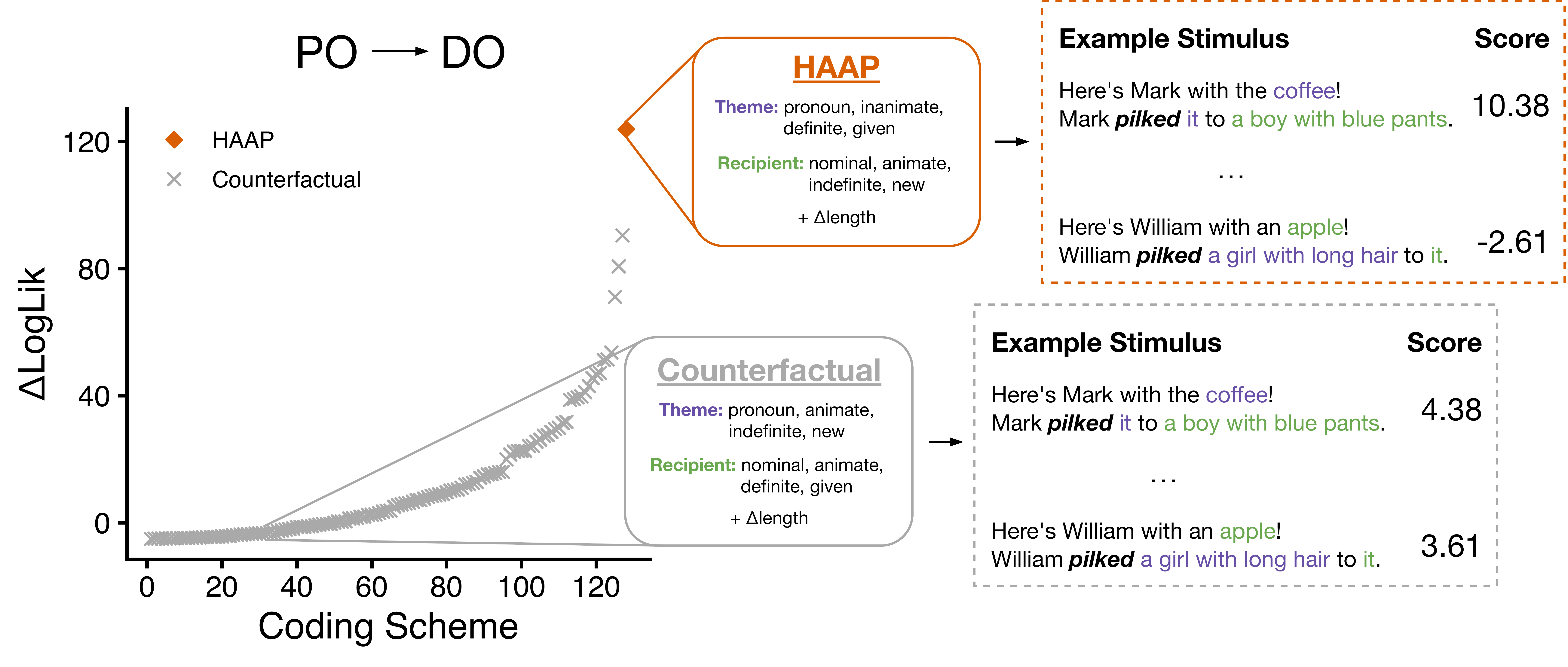}
    \end{subfigure}
    \hfill
    \begin{subfigure}[t]{0.28\textwidth}
    \centering
        \caption{\label{fig:po-do-lmer-haap}}
        \includegraphics[height=5cm]{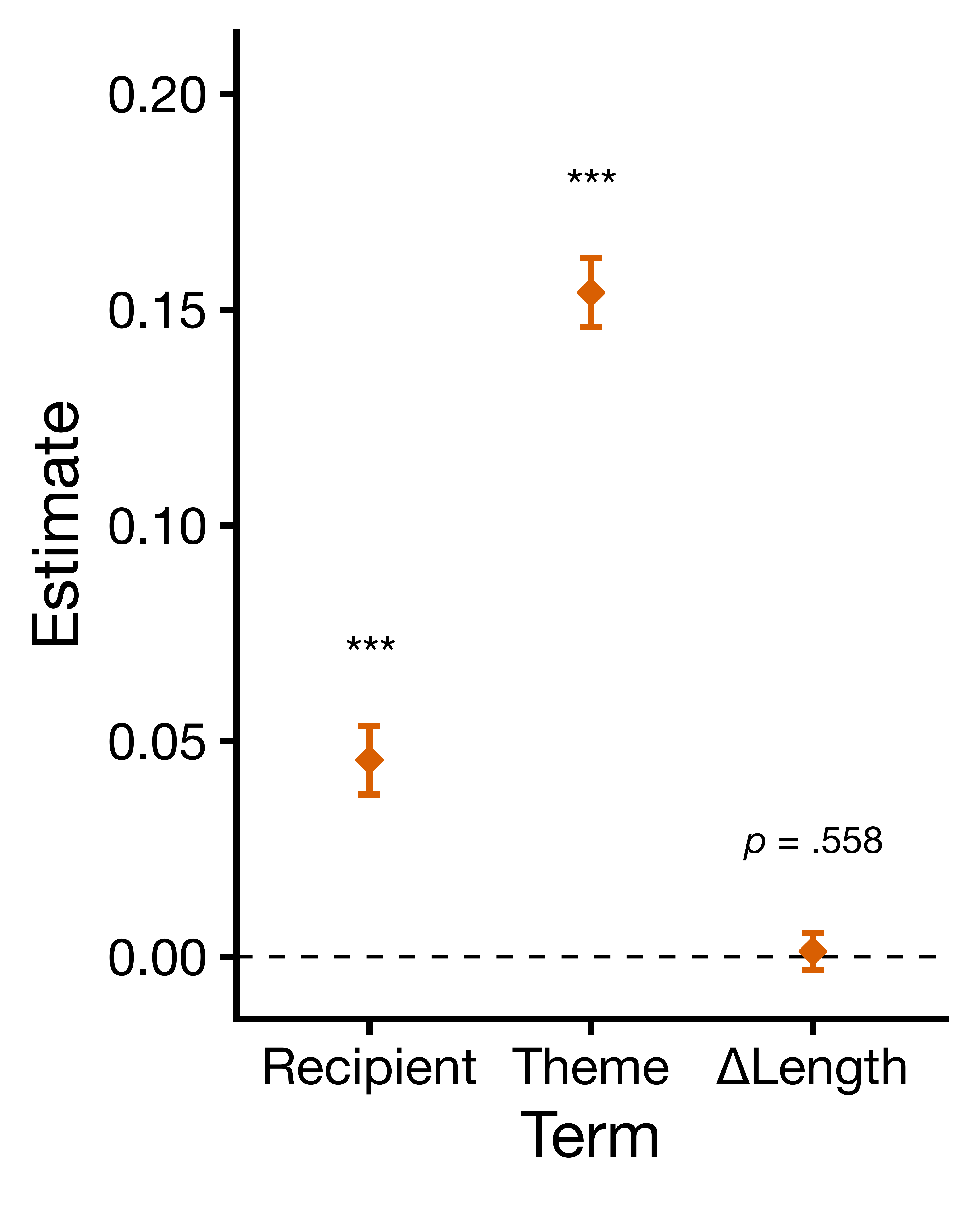}
    \end{subfigure}
    \caption{Results and analyses from our simulations. 
    \textbf{\textsf{a.}} Comparison of coding schemes in terms of the overall model fits obtained in the DO $\rightarrow$ PO generalization simulation. Scores derived using the HAAP coding scheme result in the best model fit. Example exposure stimuli and scores (recipient + theme + $\Delta$length) provided for the HAAP scheme vs. a randomly selected counterfactual scheme. 
    \textbf{\textsf{b.}} Estimated association of recipient, theme, and length-different scores under the HAAP coding scheme, at predicting LM learners' generalization of a novel verb's usage from a \textbf{DO exposure to the PO construction}. 
    \textbf{\textsf{c.}} Comparison of coding schemes in terms of the overall model fits obtained in the PO $\rightarrow$ DO generalization simulation. Scores derived using the HAAP coding scheme result in the best model fit. Example exposure stimuli and scores (recipient + theme + $\Delta$length) provided for the HAAP scheme vs. a randomly selected counterfactual scheme.
    \textbf{\textsf{d.}} Estimated association of recipient, theme, and length-different scores under the HAAP coding scheme, at predicting LM learners' generalization of a novel verb's usage from a \textbf{PO exposure to the DO construction}. 
    For both \textbf{\textsf{b}} and \textbf{\textsf{d}}, scores of the first postverbal argument of the exposure construction (recipient for DO and theme for PO) show stronger association with the LMs learners' generalization behavior.
    Model fits are estimated using the difference in log-likelihood of the fitted model relative to a null model (only random effects).
    Statistical significance calculated using a likelihood ratio test. $^{***}$: $p$ \textless{} .001.}
    \label{fig:experiment}
\end{figure}

$\Delta$length will be positive for a DO exposure whose recipient is shorter than theme, and for a PO exposure whose theme is shorter than the recipient, in accordance with harmonic alignment effects \citep{arnold2000heaviness, bresnan2007predicting}. As an example, consider the following two stimuli in the DO exposure, both of which are also shown in \Cref{fig:do-po-comparison}:

\ex. \label{ex:haap}
\a. \label{ex:haap-good} Here's Laura and Eve! Laura \textit{pilked} her a box with blocks.
\b. \label{ex:haap-ungood} Here's Mark with Sally! Mark \textit{pilked} a cup with water her.

In \Cref{ex:haap-good}, the recipient is pronominal (+1), animate (+1), definite (+1), and given (+1), giving us a recipient score of 4. The theme is nominal (+1), inanimate (+1), indefinite (+1), and new (+1), giving us a theme score of 4. The length difference between the recipient and theme is 3, which means $\Delta$length is $\log{(3)}+1$ =  2.09. This gives us a full HAAP score of 10.09. By contrast, the recipient in \Cref{ex:haap-ungood} is nominal (0), inanimate (0), indefinite (0), and new (0), and its theme is pronominal (0), animate (0), definite (0), and given (0), with the difference in lengths being -3. This gives us a recipient score of 0, theme score of 0, $\Delta$length of -2.09, and full HAAP score of -2.09.

Overall, harmonic alignment and restrictions on argument animacy allow us to fill an important gap in the literature on cross-dative generalization in two main ways. First, these effects have only been observed in the context of production data \citep{bresnan2007predicting, de2012statistical}, specifically for known verbs---their status and impact on the acquisition and generalization of partially observed, novel dative verbs is largely unknown. Second, they offer a unified framework to jointly interpret and measure the effects of changes in the feature configuration on a learner's generalization behavior. 

The \haap{} scoring scheme makes theoretical commitments to harmonic alignment and argument animacy restrictions as an explanation for cross-dative generalization. However, there are many different possible ways to assign scores to features. Specifically, it is only one out of 128 different ways to code our exposure feature configurations (see Supplemental Note 1 for the derivation of the number of scoring schemes), assuming the same format of scoring for each feature.
For instance, in one counterfactual scheme, recipients are preferred to be pronominal, animate, definite, and given while themes are preferred to be nominal, animate, definite, and given. Under this scheme, \cref{ex:haap-good} and \cref{ex:haap-ungood} are given a full score of 5.09 and 2.09, respectively (cf. 10.09 and -2.09 according to \haap{}). In addition to \haap{}, we exhaustively evaluate all possible counterfactual scoring schemes, allowing room for \textit{bottom-up} discovery of hypotheses (i.e., a scenario where a non-\haap{} scoring scheme explains model behaviors the best), as well as verification of a \textit{top-down}, human scientist-generated hypothesis (i.e., \haap{} explains model behaviors the best). 

We compare how well the scores derived using the 128 different schemes, including \haap{}, fit the LMs' generalization behavior. We fit a linear mixed-effects model, predicting the average log probability per token of the generalization set sentences, using a fixed effect for the full score (sum of the three components), and random effects for model seed, discourse template, and item (per feature configuration). We measure the strength of fit by comparing the model with fixed effects to a null model, which only has random effects. Results from these comparisons for DO and PO exposures are shown in Figures \ref{fig:do-po-comparison} and \ref{fig:po-do-comparison}, respectively. We observe that \haap{} is considerably better than the null model, and results in the best-fitting model for both generalization experiments  compared to all counterfactual schemes.

Having established that the \haap{} scoring scheme results in the best fit to the LM generalization results, we now investigate the effects of its individual components: theme score, recipient score, and $\Delta$length. For this, we use a similar linear mixed-effects model as before, this time using separate fixed effects for these components, and use the same random effects as before (model seed, discourse template, and item within feature configuration). Results from this analysis are shown in Figures \ref{fig:do-po-lmer-haap} (DO $\rightarrow$ PO) and \ref{fig:po-do-lmer-haap} (PO $\rightarrow$ DO). For the DO $\rightarrow$ PO experiment, the recipient score ($\beta$ = 0.120, $t$ = 19.353, $p <$ .001) has a larger effect than the theme score ($\beta$ = 0.037, $t$ = 5.917, $p <$ .001).
For the PO $\rightarrow$ DO experiment, it is the theme score ($\beta$ = 0.154, $t$ = 37.785, $p <$ .001) that has a larger effect than the recipient score ($\beta$ = 0.046, $t$ = 11.197, $p <$ .001). Common to both these cases is that the features of the first postverbal argument of the exposure (recipient for DO and theme for PO) have a stronger association with the model's generalization behavior than do the features of the second postverbal argument. 
This suggests that the first postverbal argument of an exposure has a privileged status in predicting an LM learner's cross-dative generalization.

\section{Discussion}

Our goal in this paper was to use LMs as animal models to generate hypotheses about how the cross-dative generalization of a novel verb relates to the features of the exposure in which it was learned from.
In building towards this goal, we first showed that LMs trained on child-directed speech can capture key generalization patterns of both known and novel dative verbs.
Our models were successfully able to predict the alternation preferences of verbs that did not alternate in the training set---i.e., they learn that \textit{kick} is more likely to occur in the DO than is \textit{explain} even if they have experienced both of these verbs \textit{only} in the PO construction.
Our models also replicated patterns of cross-dative generalization of novel verbs  shown by children in laboratory settings \citep{conwell2007early, rowland2010role, arunachalam2017preschoolers}---i.e., their cross-dative generalization was asymmetric in the same way as children, they showed advantage of cross-structure training, and they were insensitive to theme animacy when learning the verb from DO exposures. Collectively, this suggests that our models are well-positioned to be used as tools that can allow us to go beyond known results, and produce new insights about how exposures constrain cross-dative generalization of novel verbs.

To this end, the first main finding of our simulations is a statistical relationship between models' cross-dative generalization and the extent to which feature configuration of the exposure conforms to harmonic alignment effects and argument animacy restrictions (as measured by \haap{}). By harmonic alignment effects, we mean the way in which the features of the arguments (theme and recipient) were linearly organized---e.g., we expected pronominal arguments to precede nominal arguments, definite arguments to precede indefinite arguments, etc. In our results, taking the PO to DO experiments as an example, models exposed to the novel verb in the PO dative with a harmonically aligned organization of themes and recipients were more likely to generalize the verb to the DO than if they were not. It turned out the top-down, human scientist-proposed hypothesis (\haap{}) explained the model behavior the best, but our testing of counterfactual scoring schemes (the 127 non-\haap{} coding) facilitates bottom-up generation of hypotheses, had it been the case that there was a coding scheme that explained the model behavior better than the top-down hypothesis. 

From the point of view of what feature combinations \textit{restrict} generalization, the observed effects are characteristic of \textit{statistical preemption} \citep{goldberg1995constructions, goldberg2011corpus, goldberg2016partial}. Under this account, indirect evidence against a construction $K$ arises from repeatedly observing a near-synonymous construction $N$ in contexts where the learner expects $K$ to occur. 
In such a scenario, $N$ preempts the usage of $K$. Preemption has a more well-established explanatory role in literature on productive generalization in humans \citep{boyd2011learning, ambridge2012roles, ambridge2015preemption}, especially in terms of morphological phenomena \citep{aronoff1976word, kiparsky1982lexical}---e.g., observing \textit{went} preempts the usage of \textit{goed}. In the context of cross-dative generalization, taking PO to DO as an example, the preemption account suggests the following:
The usage of a novel verb in the DO construction is preempted if it is observed in a PO construction in contexts where a DO construction is expected. This is a type of \textit{counterfactual inference}: had DO construction been licensed for the novel verb, DO would have been used (because the contextual features lead to an expectation for DO), but PO was used instead. Therefore, it is less likely that DO is licensed for the novel verb. 
Harmonic alignment effects allow us to explicitly operationalize how the feature configuration of the exposure gives rise to an the expectation for a dative construction. This is especially supported by previous work---features that participate in harmonic alignment effects are strongly predictive of the choice of construction (DO vs. PO) in the production data of both children and adults \citep{bresnan2007predicting,de2012statistical}.
For example, if the recipient is short, pronominal, and definite, and the theme is long, nominal, and indefinite, then the learner might very likely expect the verb to be used in the DO. Our results show that PO exposure to a novel verb with DO-expecting features is associated with overall lower DO generalization preference, which is indicative of preemption as discussed above.

The second salient finding in our simulations was the relative effects of the features of the first and second postverbal argument. For DO exposures, the recipient is the first postverbal argument, and its \haap{} score had relatively greater effects in predicting the LM learners' generalization behavior than did the theme's \haap{} score. The opposite was true for PO exposures, with the theme (first postverbal for PO) features showing greater effects than the recipient features. This demonstrates an interesting asymmetry in the effect of general quantitative harmonic alignment between discourse and positional scales---the extent to which the first postverbal argument conforms to harmonic alignment (and animacy restrictions) has disproportionately stronger association with the learner's generalization behavior than does the second postverbal argument. 

\subsection{Novel Hypotheses for Human Experiments}

Taken together, our results lead to two novel hypotheses about the acquisition of cross-dative generalization that can be tested in the lab with children:

\begin{enumerate}
    \item Cross-dative generalization is facilitated/preempted based on the harmonic alignment of feature configurations and argument positions in the exposure context.
    \item There is a discrepancy between the effect of first vs. second postverbal argument features, where the former has a more noteworthy effect on cross-dative generalization.
\end{enumerate}

\noindent The first hypothesis concerns a more general type of facilitatory/preemptive effects licensed by the feature configurations of the exposure context. The second hypothesis targets the specific privileged status of the first postverbal argument of the exposure construction, which was shown to have a substantially greater effect of cross-dative generalization in our experiments than the second post-verbal argument. Both these hypotheses go beyond the existing role of \textit{discourse and positional prominence}, which has so far largely focused on processing and comprehension advantages \citep[\textit{i.a.}]{birch1995effect,birch2000syntactic,foraker2007role,arnold2008put,kember2021processing}, and not on acquisition.

We sketch out two experiments that can allow the testing of these hypotheses with children, primarily focusing on the design of the experimental stimuli.
The high level experimental setting we propose is similar to the experiments we have conducted on LM learners, as well those conducted with children in previous work \citep{gropen1989learnability, conwell2007early, arunachalam2017preschoolers}: the learner is exposed to a novel verb in a given dative construction and is then tested on their usage of the novel verb in the unmodeled dative construction. 
One possible test is the comprehension task used by \citet{arunachalam2017preschoolers}, where participants are asked to identify the scene (given two choices) that describes the event in a sentence (in our case, a sentence where the novel verb appears in an unmodeled dative construction).\footnote{An alternative experimental paradigm to test our proposed hypotheses would be to use a production task as in \citet{gropen1989learnability} and \citet{conwell2007early}. However, such production tasks suffer from general problems associated with free-form response elicitation, where the target production is difficult to control for due to high variation across individuals and trials. Since our discussion in this paper only focuses on experimental hypotheses, we leave precise decisions about the generalization task to future work in collaboration with language acquisition researchers.}
Generalization strength can then be quantified by measuring the proportion of time the right scene was chosen by the learners. Alternatively, we could adopt the acceptability judgment paradigm which has been shown to be successfully applicable to children \citep{ambridge2011assessing} using cartoon-face versions of Likert scales, which is a more direct translation of the simulation paradigm. 

\begin{table}[!t]
\centering
\begin{tabular}{@{}lllll@{}}
\toprule
\textbf{Experiment} & \textbf{Generalization} & \textbf{\begin{tabular}[c]{@{}l@{}}First postverbal\\ arg. of exposure\end{tabular}} & \textbf{Theme} & \textbf{Recipient} \\ \midrule
1a & DO $\rightarrow$ PO & Recipient & \haapmax{} & \haapmax{} \\
1b & DO $\rightarrow$ PO & Recipient & \haapmin{} & \haapmin{} \\
1c & PO $\rightarrow$ DO & Theme & \haapmax & \haapmax{} \\
1d & PO $\rightarrow$ DO & Theme & \haapmin & \haapmin \\ \midrule
2a & DO $\rightarrow$ PO & Recipient & \haapmin & \haapmax \\
2b & DO $\rightarrow$ PO & Recipient & \haapmax & \haapmin \\
2c & PO $\rightarrow$ DO & Theme & \haapmax & \haapmin \\
2d & PO $\rightarrow$ DO & Theme & \haapmin & \haapmax \\ \bottomrule
\end{tabular}%
\caption{Feature configuration setup for our proposed experiments. \haapmax{} indicates the features are maximally compliant with harmonic alignment and animacy prototypicality effects as quantified in our study, whereas \haapmin{} indicates minimal compliance.}
\label{tab:proposed-experiments}
\end{table}

\begin{table}[]
\centering
\resizebox{\textwidth}{!}{%
\begin{tabular}{@{}llll@{}}
\toprule
\textbf{Construction} & \textbf{Argument} & \textbf{\haapmax{}} & \textbf{\haapmin{}} \\ \midrule
\multirow{2}{*}{DO} & Theme & \textit{\begin{tabular}[c]{@{}l@{}}a ball with stars, a cup with no lid, a cookie with\\ chocolate frosting, a glass with milk, ...\end{tabular}} & \textit{\begin{tabular}[c]{@{}l@{}}it, them (inanimate, disambiguated from the \\ previous sentence)\end{tabular}} \\ \cmidrule(l){2-4} 
 & Recipient & \textit{me, him, her, them, us} & \textit{\begin{tabular}[c]{@{}l@{}}a girl with green hair, a boy with round glasses,\\ a girl with blue hat, a horse with the saddle, ...\end{tabular}} \\ \midrule
\multirow{2}{*}{PO} & Theme & \textit{\begin{tabular}[c]{@{}l@{}}it, them (inanimate, disambiguated from the\\ previous sentence)\end{tabular}} & \textit{\begin{tabular}[c]{@{}l@{}}a ball with stars, a cup with no lid, a cookie with\\ chocolate frosting, a glass with milk, ...\end{tabular}} \\ \cmidrule(l){2-4} 
 & Recipient & \textit{\begin{tabular}[c]{@{}l@{}}a girl with long hair, a boy with blue pants, \\ a bear with an apple, a boy with round glasses, ...\end{tabular}} & \textit{me, him, her, them, us} \\ \bottomrule
\end{tabular}%
}
\caption{\revision{Stimuli design for theme and recipient arguments in the proposed experiments, spanning from those that are maximally and minimally compliant with \haap{}.}}
\label{tab:example-stimuli}
\end{table}

In terms of stimuli design, we propose selecting arguments of the exposure constructions that either maximally or minimally conform to our \haap{} scores in order to verify our hypotheses. Note that the lexical items themselves do not necessarily have to be restricted to the ones we used, since the scoring is based on features and not the surface forms themselves. Insofar as our hypotheses hold, we should expect maximal effects of preemption or facilitation of generalization with our maximal/minimal design. In the first experiment, we propose a 2$\times$2 design, where the two types of exposure constructions (DO/PO) will be paired with two types of argument feature choices---both being either maximally \haap{}-compliant (\haapmax{}) or minimally \haap{}-compliant (\haapmin{}). This will allow us to test whether harmonic alignment effects, as found in our simulations, hold for children's cross-generalization of a novel dative verb (the first hypothesis). In the second experiment, we will instead focus on the apparent privileged role of the first postverbal argument of the exposure condition. This too will be a 2$\times$2 design, with the exposure conditions being the same (DO/PO) but the argument feature choices varying in terms of whether or not the first postverbal argument of the exposure is \haap{}-compliant. Here, if one argument is \haapmax{}, then we deliberately make the other \haapmin{}. This is so that if the first-postverbal alone is sufficient enough to determine the degree of cross-dative generalization, then we should observe greater cross-dative generalization when the first postverbal argument is \haapmax{}. 

\Cref{tab:proposed-experiments} shows all unique experimental conditions along with the choice of arguments (with respect to \haapmax{} vs. \haapmin{}), \revision{while \Cref{tab:example-stimuli} shows a sample of theme and recipient items conforming to the various conditions}. Based on our hypotheses, we expect cross-dative generalization to pattern as follows: If hypothesis 1 is holds for children, then we expect them to generalize more in 1a and 1c (both arguments \haapmax{}) than in 1b and 1d (both arguments \haapmin{}), respectively for DO and PO exposures. For hypothesis 2 to hold, we should find children to show better generalization in 2a and 2c (First postverbal argument \haapmax{}) than in 2b and 2d (First postverbal argument \haapmin{}), respectively for DO and PO exposures.

If these hypotheses indeed hold in the lab, then it would mean that children are sensitive to the preemptive effects of argument feature configurations. Preemption based accounts for cross-dative generalization of novel verbs have been underexplored, presumably due to the expectation that such an investigation would require many exposures to the alternate form both at the type and token levels to serve as sufficient indirect evidence. This can be difficult due to the large space of contextual features involved, especially in studies involving children. However, our simulations demonstrate that preemption-compatible effects can be fruitfully investigated even with relatively limited, single-sentence exposures (cf. \citep{goldberg2011corpus}). If such effects are also found in studies of children, this can potentially motivate a wider range of studies with
similar design that test preemption effects in the lab via novel word learning.

\section{Conclusion}
By proposing a general framework of hypothesis generation and instantiating it with a particular research problem, we were able to put forward hypotheses about the role of harmonic alignment in the acquisition context, which has never been investigated in prior work. Furthermore, to the best of our knowledge, this work is the first instance of experimental hypothesis generation in the domain of language development under a systematic framework, and the first to derive hypotheses of this specificity. Taken together, our work makes both a domain-specific contribution (specific verb acquisition and generalization hypotheses to test in the lab) and a domain-general contribution (a framework for systematic hypothesis generation using simulated learners). 

\revision{We see our framework's functional role to be similar to the role of computational modeling in theory building and testing discussed in \citet{norris2005computational,mcclelland2009place,guest2021computational}, among others. While our interpretation of the models studied is different from computational models, our framework also forces researchers to spell out commitments about preconditions, simulation details, and translation to new hypotheses. Furthermore, the simulations allow for a larger systematic search of the hypothesis space, combining hypothesized features in the literature, which may not be feasible with human subjects due to practical constraints. This allows us to discover unexpected interactions or effects that are obscured in single-factor studies. Therefore, our framework serves a functionally similar role in science of surfacing consequences that existing approaches may miss, thereby contributing to theoretical development.}

\section{Limitations and Future Directions}
\label{sec:limitations}

\subsection{Mechanistic and developmental differences between LMs and humans}
Our work does not make any claims about the mechanistic similarities between how children and LMs carry out cross-dative generalization. It is possible that observations about LMs replicating known patterns of cross-dative generalization are subject to multiple realizability \citep{quine1951main, fodor1988connectionism}---two systems with completely different processing mechanisms may show similar input-output behaviors \citep[cf.][]{cao2024explanatory}. Further investigations should be made in order to make any claims about processing-level similarity, which is not in the scope of the current work.

\revision{Additionally, our current work draws conclusions at the level of the LMs' training outcomes, rather than from their learning trajectories. It is possible that while the end state (i.e., checkpoint during training) of our models can achieve high fidelity results in replicating known patterns of cross-dative generalization, their developmental trajectory might look different to humans \citep{chang2022word, evanson2023language}. Nonetheless, our study highlights how interesting insights can be derived from outcome-aligned models (i.e., ones that end up achieving qualitatively similar generalization patterns as found in previous human studies). We leave a thorough analyses of models' developmental trajectories and their alignment with human learning development to future work.}

\subsection{Phenomenon detection using \texttt{spacy}} \revision{Another limitation of our work is that we rely on the \texttt{spacy} library \citep{spacy} to aid in the detection of dative constructions as well as in selecting our verbhood examples. We therefore inherit any limitation that is incurred by \texttt{spacy}--e.g., parsing and tagging errors. Future work could utilize more recent libraries like \texttt{UD-English-CHILDES} \citep{yang-etal-2025-ud} and \texttt{CAIT} \citep{padovani-etal-2026-cait} that are specialized for child-directed speech.}

\subsection{Effect of distributional cues from a broader set of exposures}
The experiments in this work presented learners with controlled exposure stimuli during training. This design and the choice of contextual features were motivated by nonce word learning research conducted in the lab. However, the learning of novel verbs in the wild is likely affected by a much broader set of distributional cues that go beyond the range covered by our experiments. Especially in the case of dative verbs, many ditransitive verbs also have transitive uses (e.g., \textit{I \textbf{kicked} the ball, She \textbf{kicked} him on the knees}). What is the effect of such encounters on the learning of the double object dative uses of \textit{kicked}? More broadly, do children generalize from \textit{any} type of indirect evidence about verb use, or is their generalization mostly dependent on encounters with constructions involving the same set of arguments? These questions are central to debates surrounding preemption and entrenchment in the acquisition of phrasal constructions \citep{stefanowitsch2008negative, goldberg2011corpus, boyd2011learning, ambridge2012roles, ambridge2015preemption}. While recent work has shown LMs to also generalize from indirect evidence \citep{wei-etal-2021-frequency, jumelet-etal-2021-language, misra-mahowald-2023-rare, patil2024filtered, yao2025both, leong2026manipulating, wang2026disentangling}, questions about how indirect evidence modulates productive generalization and its interplay with the feature configurations of the relevant exposures are still open. 
This could also be contrasted with more explicit signals of grammar learning feedback \citep{nikolaus2023communicative, nikolaus2026modelling}.
We hope our method and general framework presented here can help investigate these questions in both language models and humans in future work.

\subsection{Generalization of Framework}
\revision{We discuss a few directions in which our framework can be generalized to other phenomena to illustrate its broader applicability. One example is to explore the hypothesis space of the factors responsible for the acquisition of factivity. For instance, verbs such as \textit{know} presuppose the truth of their complement, whereas verbs like \textit{think} do not. We can investigate the properties of the input that lead to this facitivity difference by systematically manipulating syntactic distribution and the discourse contexts of the verbs, drawing from work such as \citet{dudley2017role}, to examine their effects on the learning outcome. Similarly to our work on dative alternation, exploring the combinations of these factors could reveal interactions that are not evident from independent considerations of these factors, motivating targeted hypotheses for follow-up human experiments.}

\revision{Another way the proposed framework can be adapted is in the context of understanding the cues to semantic categories. For instance, it has long been hypothesized that category membership could partially emerge from information present in a child’s linguistic environment \citep{bowerman1980structure, waxman1995words, borovsky2006language, wilson2023conceptual}
This knowledge could arise not only from explicit (\textit{a goat is an animal}) but also implicit cues---e.g., ``\textit{Which animal did you mean? Did you mean that goat over there?}'' \citep{wilson2023conceptual}. 
Our framework can be used to quantify the contribution of such cues, where we can perform a set of novel word learning trials with LMs, and manipulate the extent to which the LM learners observe certain kinds of cues. For instance, one could come up with explicit and implicit cues with a known category (e.g., \textit{bird}) and a novel noun (e.g., \textit{wug}) that is posited as a member of the category, and test the extent to which the model establishes a hypernymy relation between the two as a function of the type and amount of cues provided (\textit{X is a Y, Xs and other Ys, Ys such as Xs, Which Y did you mean? Did you mean this X over here?,} etc.)}

\section{Code and Data Availability}
All our code and data, which include python scripts for training models, data generation, and simulations, as well as raw stimuli files, and results from simulations are available on our github: \texttt{Anonymized URL}. Details of every statistical analysis is provided in Section 6 of the Supplementary Materials.

\bibliographystyle{apalike}
\bibliography{refs}

\appendix

\section{Language Model Hyperparameter and Implementation Details}

\Cref{tab:trainingdetails} shows the full set of training details of the LM learners used in this work. To train LMs from scratch, we used the the \texttt{transformers}\footnote{\url{https://github.com/huggingface/transformers}} library \citep{wolf-etal-2020-transformers}. 
We trained 5 different instances of an LM using the aforementioned hyperparameters, each using a different training seed. All our LMs can be found on the Huggingface Hub using the url: \texttt{(anonymized)}.
We compute log probabilities from these models using the \texttt{minicons} library \citep{misra2022minicons}.

\begin{table}[!ht]
\centering
\begin{tabular}{@{}lr@{}}
\toprule
\textbf{(Hyper)parameter} & \textbf{Value} \\ \midrule
Architecture & OPT \citep{zhang2022opt} \\
Embed size & 256 \\
FFN dimension & 1,024 \\
Num. layers & 8 \\
Attention heads & 8 \\
Vocab size & 8,192 \\
Max. seq. length & 128 \\
Batch size & 16 \\
Final learning rate & 0.003 \\
Learning rate scheduler & Linear \\
Warmup steps & 24,000 \\
Epochs & 10 \\
Training data & AO-CHILDES \citep{huebner2021using} \\
Total parameters & 8.3M \\
Training size & 4.71M tokens \\
Compute & 1x NVIDIA RTX 6000 Ada \\\bottomrule
\end{tabular}%
\caption{LM Training details}
\label{tab:trainingdetails}
\end{table}

\section{Stimuli Generation}

This section presents our pipeline for stimuli generation, for both the main simulations, as well as the precondition experiment that tests whether our LM learners replicate the findings of \citep{arunachalam2017preschoolers}.

\subsection{Main Simulations}

The stimuli in our main simulations consist of utterances in the DO and PO dative constructions. The utterances themselves primarily vary in terms of the features of the theme and recipient, since a core difference between the  constructions is the ordering of these arguments, and that features of these two arguments form the hypothesis space we tackle in this paper. We specifically focus on five kinds of features, each for the theme and recipient: \textsc{Pronominality, Animacy, Definiteness, Length,} and \textsc{Discourse Givenness}. Each feature is associated with a particular set of lexical items (e.g., \textit{he/she/etc.} if the \textsc{Pronominality} feature is `\textit{pronoun}'), and the final set of lexical items for an argument (theme/recipient) for a given instance in our stimuli is defined by a particular feature combination. Below we discuss each feature, along with a brief description of the representative lexical items that we used in our stimuli.

\paragraph{\textsc{Pronominality}}
Pronominality (pronominal vs. nominal) has been shown to have a significant effect on relative acceptability and preference between dative constructions \citep{bresnan2007predicting, de2012statistical}.
In novel dative verb learning studies, recent work has found pronominal recipients (especially \textit{me}, a frequent recipient in child-directed speech) to facilitate DO comprehension \citep{conwell2019effects}.
In our stimuli, we included 8 different pronouns, accounting for variation in animacy (\textit{him/her} vs. \textit{it}) and definiteness (\textit{her} vs. \textit{someone}). 
For non-pronominal items, we used 18 different noun items (not accounting for any modification) -- e.g., \textit{mommy, daddy, cat, dog, bear, cookie, book, ball, lego, chair}, etc.

\paragraph{\textsc{Animacy}}
The animacy of the theme and recipient has been centrally discussed in theoretical and experimental literature surrounding dative alternation \citep[\textit{i.a.}]{gropen1989learnability, bresnan2007predicting, hovav2008english, beavers2011aspectual, de2012statistical, arunachalam2017preschoolers}. 
In experimental work, the combination of animate theme and animate recipient in a DO construction has been found to be difficult to comprehend for children \citep{rowland2010role, arunachalam2017preschoolers}.
In our stimuli, we included 27 different entries for animate ($N$ = 14) and inanimate ($N$ = 13) items, accounting for variation in pronominality (\textit{her} vs. \textit{it} for pronominal; \textit{mommy} vs. \textit{book} for nominal), but not accounting for any modification (determiner, adjectival, prepositional), which is more relevant to definiteness and length.

\paragraph{\textsc{Definiteness}} Definiteness is inextricably linked to the discourse status of an item---definite items (\textit{the ball}) are often discourse given, while indefinite items (\textit{a ball}) are often used to introduce new discourse entities.
The discourse status of arguments have been found to affect the choice between dative constructions, at least in adults \citep{bresnan2007predicting}---definite items (usually discourse given) tend to occur before indefinite items \citep{arnold2000heaviness, wasow2002postverbal}.
This explains the preference for DO when the recipient is definite, and preference for PO when the theme is definite.
Our stimuli accounts for definiteness (definite vs. indefinite) via different pronouns \textit{(him, her, them, it} vs. \textit{someone, something}), proper nouns (\textit{mommy, daddy}), and determiner modification to the set of nouns (\textit{the ball} vs. \textit{a ball}).

\paragraph{\textsc{Length}} The length of the arguments has also been shown to play a role in postverbal word order in English \citep{aissen1999markedness, arnold2000heaviness, wasow2002postverbal}. Heavy (more complex, longer) phrases tend to occur after lighter (less complex, shorter) phrases, with this observation dating back to \citet{behaghel1909beziehungen} (noted by \citealt{arnold2000heaviness}).
This pattern is also reflected in dative alternation: themes tend to be longer than recipients in DOs, while the opposite is true for POs \citep{bresnan2007predicting, de2012statistical}.
We measured length in terms of number of words, and treated it as a continuous value, measured in terms of the difference in the number of words in the theme vs. recipient, following \citet{bresnan2007predicting} and \citet{de2012statistical}. We varied length by adding adjectival and prepositional modification, e.g., \textit{the ball} $\rightarrow$ \textit{the red ball} for adjectival modification, or \textit{the ball} $\rightarrow$ \textit{the ball with a star on it} for prepositional modification. We vary the length difference to be all integers in the interval [-6, 6].

\paragraph{\textsc{Discourse Givenness}} The final feature we considered in our hypothesis space is discourse givenness; prior work has observed that given information is typically mentioned before new information \citep{clark1977psychology, gundel1988universals, arnold2000heaviness}.
Aligned with this observation, corpus analyses of child-directed speech as well as adult conversations show that the DO construction is preferred when the recipient is given, while PO is preferred when themes are given \citep{bresnan2007predicting, de2012statistical}.
This is also borne out in experimental evidence---both children and adults showed the \textit{given-before-new} order in their production of dative constructions \citep{stephens2015dative}. To specify givenness, we inserted a sentence before the main dative stimulus, which contains the agent and the information that is given.\footnote{Our inclusion of the agent follows multiple novel dative verb learning studies that familiarized child learners with the agent and at least one of the arguments \citep{conwell2007early, rowland2010role, arunachalam2017preschoolers, conwell2019effects}.}
We used three different variations for our givenness-specification templates, while also varying what argument in the dative construction was \textit{given}. For the latter, we considered three options: 1) only the agent; 2) the agent and one of the arguments (which gave us two different conditions); 3) the agent and both of the argument (for which we counterbalanced the order of theme and recipients for each instance in this condition). The templates across these conditions are given in \Cref{tab:givenness-templates}.

\begin{table}[]
\centering
\resizebox{0.8\textwidth}{!}{%
\begin{tabular}{@{}llll@{}}
\toprule
\textbf{Given Condition} & \textbf{Template 1} & \textbf{Template 2} & \textbf{Template 3} \\ \midrule
Agent & Do you see \texttt{\{agent\}}? & Look, it's \texttt{\{agent\}}! & Here's \texttt{\{agent\}}! \\
\addlinespace
Agent + One Argument & \begin{tabular}[c]{@{}l@{}}Do you see \texttt{\{agent\}}\\ and \texttt{\{given-arg\}}?\end{tabular} & \begin{tabular}[c]{@{}l@{}}Look, it's \texttt{\{agent\}} \\ with \texttt{\{given-arg\}}!\end{tabular} & \begin{tabular}[c]{@{}l@{}}Here's \texttt{\{agent\}} \\ with \texttt{\{given-arg\}}!\end{tabular} \\
\addlinespace
Agent + Both Arguments & \begin{tabular}[c]{@{}l@{}}Do you see \texttt{\{agent\}} \\ and \texttt{\{given-arg1\}} \\ and \texttt{\{given-arg2\}}?\end{tabular} & \begin{tabular}[c]{@{}l@{}}Look, it's \texttt{\{agent\}} \\ and \texttt{\{given-arg1\}} \\ and \texttt{\{given-arg2\}}!\end{tabular} & \begin{tabular}[c]{@{}l@{}}Here's \texttt{\{agent\}} \\ with \texttt{\{given-arg1\}}\\ and \texttt{\{given-arg2\}}!\end{tabular} \\ \bottomrule
\end{tabular}%
}
\caption{Givenness templates across different conditions. \texttt{\{agent\}} and \texttt{\{given-arg\}} represent the agent and the argument that is given, respectively.}
\label{tab:givenness-templates}
\end{table}

\begin{table}[]
\centering
\begin{tabular}{@{}lll@{}}
\toprule
\textbf{Feature}               & \textbf{Value} & \textbf{Examples}                                                  \\ \midrule
\multirow{2}{*}{Animacy}       & Animate        & \textit{mommy, daddy, dog, boy, girl, horse, bear, horse}          \\
                               & Inanimate      & \textit{cup, ball, book, food, cookies, lego, chair, box}          \\ \midrule
\multirow{2}{*}{Pronominality} & Pronominal     & \textit{him, her, it, them, us, me, someone, something}            \\
                               & Nominal        & \textit{mommy, daddy, dog, boy, girl, book, cup, ball}             \\ \midrule
\multirow{2}{*}{Definiteness}  & Definite       & \textit{the ball, the book with pictures, the boy with blue pants} \\
                               & Indefinite     & \textit{a ball, a book, a boy, a lego, a chair, a box with blocks} \\ \bottomrule
\end{tabular}
\caption{\revision{Example items across Animacy, Pronominality, and Definiteness features.}}
\label{tab:example-lexical-items}
\end{table}

These considerations give us 8 binary features (4 each for the theme and recipients---\textsc{Pronominality, Animacy, Definiteness, Discourse Givenness}), and 13 possible values for \textsc{Length}, which results in 3328 possible feature configurations ($2^8 \times 13$). However, a majority of these configurations result in empty sets of lexical items---e.g., only a length difference of 0 is possible when both themes and recipients are pronominal, it is pragmatically infelicitous to have an item be given in the discourse but referred to using an indefinite article (e.g., \textit{Do you see Jenny with \textbf{the ball}? She pilked \textbf{a ball} to me.}), etc. On discarding such cases, we end up with a total of 756 possible feature configurations per dative construction, from which we sample lexical items to generate our stimuli items. \revision{\Cref{tab:example-lexical-items} shows a sampling of lexical items for the Animacy, Prononominality, and Definiteness features.\footnote{Givenness is omitted since it is dictated by whether or not the previous discourse mentions a noun, and length is omitted for brevity.}} We sample a total of 8 items for each feature configuration, using a different proper name for each sampled item as the agent (from the list: \{\textit{Laura, Mark, Sarah, William, Alex, Judy, Michael, Jenny}\}). This gives us 6,048 items per givenness template per dative, amounting to a total of 36,288 items.



\subsection{Additional stimuli for replicating the results of \citet{arunachalam2017preschoolers}}

In the experimental stimuli of the original study \citep{arunachalam2017preschoolers}, the recipients and themes were both definite, discourse given, and short (2 word noun phrases). They differed only in their animacy---with themes being inanimate and recipients being animate. However, this only corresponded to a single hypothesis configuration in our exposure stimuli, yielding a total of 16 items. Therefore, for the purpose of this analysis, we considered a larger set of animate ($N$ = 11) and inanimate ($N$ = 11) items, and sampled a total of 220 items in each condition, giving us 440 new stimuli. Finally, we used the same preceding discourse context as \citet{arunachalam2017preschoolers}---``Here's \texttt{\{agent\}}. Hi \texttt{\{agent\}}! Here's a \texttt{\{theme\}} and a \texttt{\{recipient\}}!''---but we counterbalance in the relative order of the theme and recipient mentions in the discourse context.

\section{Detecting dative constructions in AO-CHILDES}
This section describes our pipeline to automatically detect instance of the DO and the PO dative constructions. We apply this pipeline on the validation and the test set to detect verbs that do not alternate in our training set, and to create our generalization set.
We used \texttt{spacy} \citep{spacy} to extract dependency parses and parts-of-speech tags of all utterances in AO-CHILDES.
Then, we used two different heuristics for extracting verbs used in the DO and PO constructions. To detect PO constructions, we checked if the following conditions were true: (1) the preposition \textit{to} occurs in the sentence; (2) there is a direct dependency between \textit{to} and the verb; (3) there exists a direct \texttt{pobj} dependency relation between the \textit{to} in (1) and (2) and a noun/pronoun in the sentence; and (4) there exists a \texttt{dobj} dependency relation between the verb and a noun/pronoun separate from the noun/pronoun in (3).
For DO, we checked if: (1) the verb had an \texttt{iobj} path with a noun/pronoun in the sentence; and (2) the verb also had a separate \texttt{dobj} path with a separate noun/pronoun in the sentence that was linearly to the right of the noun/pronoun in (1). 
In both these cases, we restricted the verb lemmas to those that appear in the list of dative verbs in \citet{levin1993english}. Applying this pipeline on the AO-CHILDES training set yields 5261 DO utterances and 2724 PO utterances, while on the test and development sets yields 160 DO and 95 PO utterances.

\section{Evaluating Verbhood in Simulations}
\label{sec:verbhood}

The basic criterion we expected our LM learners to meet in our novel word-learning simulations was the verbhood of the novel word---they should treat the novel verb as a verb as opposed to other parts of speech.
This is the main criterion used to select the best embedding state of the novel verb in the LMs, before the test phase where all vector representations in models are frozen and generalization measures are computed.
This condition is also similar to the main analysis of \citet{kim2021testing}, where evidence of the above behavior is indicative of abstract category-based inference in the LM learners.
In this section we present results from this verification analysis, which will serve as a sanity check that the basic syntactic category of the novel tokens was actually being learned by the LMs.

Our validation contexts contain two sets of \revision{200} sentences each, one of which places the novel verb in verb-expecting contexts (\textit{You \blank{} the boat.}), and the other places it in non-verb expecting contexts (\textit{That's a \blank{} of the zoo.}). \Cref{tab:verbhood-examples} shows five examples of verb expecting and non-verb expecting sentences used as part of our verbhood verification examples.

\revision{We constructed this set using utterances from subsets of the full CHILDES corpus \citep{macwhinney2000childes} that are syntactically annotated with constituency parses and part-of-speech tags. From these utterances, we extracted sentences containing an unambiguous past-tense verb (\texttt{VBD}). We excluded verbs that occurred next to a pronoun, another verb, or sentence-final punctuation, as well as utterances containing wh-words or question marks, since these included auxiliary and copular contexts in which it was not always clear that the target position required a verb. To create non-verb expecting contexts, we followed the same procedure for nouns (\texttt{NN}) and adjectives (\texttt{JJ}). To avoid lexical overlap with our exposure stimuli, we removed any frame containing a lexical item that was used to generate exposure stimuli. We then sampled 200 sentences from the verb set, and 100 each from each non-verb set (\texttt{NN} and \texttt{JJ}), giving us 200 verb-expecting contexts and 200 non-verb-expecting contexts. For each of these sentences, we replaced the selected verb/non-verb with our novel verb.}

\begin{table}[]
\centering
\resizebox{0.8\textwidth}{!}{%
\begin{tabular}{@{}ll@{}}
\toprule
\textbf{Verb Expecting} & \textbf{Non-verb Expecting} \\ \midrule
jack \blank{} the treasure from the sleeping giant. & oh , you need the other \blank{} now . \\
he \blank{} a lot of things right now. & yeah, santa claus brought that \blank{} to you . \\
that \blank{} like a fish. & i'm forgetting \blank{} . \\
louise \blank{} that. & let's make a \blank{} . \\
look, somebody \blank{} something. & there, now nina has a \blank{}. \\ \bottomrule
\end{tabular}%
}
\caption{Example verb expecting and non-verb expecting sentences from our verbhood validation set. In practice, \blank{} is replaced with the surface form of the novel dative verb (here, \textbf{[pilked]}).}
\label{tab:verbhood-examples}
\end{table}

The target criterion we used in our novel verb learning simulations was that the average log probability of verb-expecting sentences should be greater than the log probability of non verb-expecting sentences.
We denote this difference as ``Verbhood $\Delta$''.
We use the embedding of the novel verb token at the epoch that results in the maximum Verbhood $\Delta$ as the final embedding of the novel token.
Since Verbhood $\Delta$ is a difference measure with positive values signifying greater verbhood, an LM that has made the right category-based inference should show Verbhood $\Delta$ values that are substantially greater than 0.0. We measure the average Verbhood $\Delta$ values from our main exposure experiments.
In addition, we also measure and report the ``Verbhood Accuracy'', which is the proportion of time (across all 5 model seeds) the log probability of using the novel verb token (after exposure) in verb-expecting contexts was greater than using it in non-verb-expecting contexts.
This measure is correlated with Verbhood $\Delta$ but provides a stricter measure of verbhood---an ideal LM should show very high verbhood accuracies, with a strict upper-bound of 100\%. Since this is a comparison between pairs, chance performance is 50\%.

\begin{figure}
    \centering
    \includegraphics[width=0.5\linewidth]{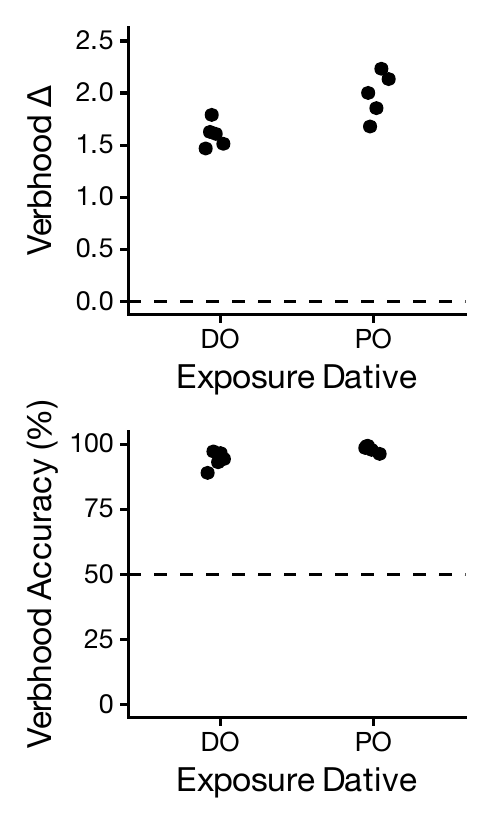}
    \caption{Average Verbhood $\Delta$s and Accuracies across different adaption dative types. Note that there is no upper/lower-bound for Verbhood $\Delta$, since they are differences in log-probabilities, and can theoretically be infinite in either direction. $\Delta$s above 0 indicate the model prefers the novel verb in verb-expecting contexts over non-verb-expecting contexts. Each point represents the result of a single LM seed.}
    \label{fig:verbhood}
\end{figure}

\Cref{fig:verbhood} shows the average Verbhood $\Delta$ values and accuracies from our 5 LM model runs for both types of exposure constructions (DO and PO). Across both exposure datives, LM learners show positive average Verbhood $\Delta$s, which are all significantly different from 0 ($p <$ .001, calculated using results across model seeds, exposure datives, and stimuli). The Verbhood Accuracies are all close to perfect, with an average of approximately 96.2\%. This suggests that our basic criterion for LMs learning to treat the novel tokens as verbs was satisfied most of the time, providing us good grounding to formulate and conduct finer-grained linguistic analyses.

\begin{table}[!t]
\centering
\resizebox{\textwidth}{!}{%
\begin{tabular}{@{}cclll@{}}
\toprule
\multirow{2}{*}{\textbf{Alternation Behavior}} &
  \multirow{2}{*}{\textbf{Dative}} &
  \multirow{2}{*}{\textbf{Lemma (Freq. in Training)}} &
  \multicolumn{2}{c}{\textbf{Example Test sentences}} \\ \cmidrule(l){4-5} 
 &
   &
   &
  \multicolumn{1}{c}{\textbf{DO}} &
  \multicolumn{1}{c}{\textbf{PO}} \\ \midrule
\multirow{6}{*}{\begin{tabular}[c]{@{}c@{}}
Non-alternating in\\training but \\actually alternating\\(\textsc{naba})
\end{tabular}} &
  DO &
  {\begin{tabular}[c]{@{}l@{}}assign (1), guarantee (1), owe (7), \\ promise (1), rent (2), trade (5)\end{tabular}} &
  {\begin{tabular}[c]{@{}l@{}}Nina \textbf{assigned} him a task\\ I \textbf{owed} them the full amount\\ The boy \textbf{promised} the girl a meal\\ Ryan \textbf{rented} us some books\end{tabular}} &
  {\begin{tabular}[c]{@{}l@{}}Nina \textbf{assigned} a task to him\\ I \textbf{owed} the full amount to them\\ The boy \textbf{promised} a meal to the girl\\ Ryan \textbf{rented} some books to us\end{tabular}} \\ \cmidrule(l){2-5} 
 &
  PO &
  {\begin{tabular}[c]{@{}l@{}}bat (1), bounce (5), deliver (7), \\ drag (1), haul (2), kick (3)\end{tabular}} &
  {\begin{tabular}[c]{@{}l@{}}Laura \textbf{bounced} him a round toy\\ Ethan \textbf{delivered} us some boxes\\ Nina \textbf{hauled} mommy the furniture\\ John \textbf{kicked} a stranger an egg\end{tabular}} &
  {\begin{tabular}[c]{@{}l@{}}Laura \textbf{bounced} the round toy to him\\ Ethan \textbf{delivered} some boxes to us\\ Nina \textbf{hauled} the furniture to mommy\\ John \textbf{kicked} an egg to a stranger\end{tabular}} \\ \midrule
\multirow{6}{*}{\begin{tabular}[c]{@{}c@{}}
Non-alternating in\\training and\\actually non-alternating\\(\textsc{nana})
\end{tabular}} &
  DO &
  {cost (5), wish (6)} &
  {\begin{tabular}[c]{@{}l@{}}John \textbf{cost} him some money\\ Ryan \textbf{cost} everyone 2 cents\\ Lily \textbf{wished} him good health\\ You \textbf{wished} us a good weekend\end{tabular}} &
  {\begin{tabular}[c]{@{}l@{}}John \textbf{cost} some money to him\\ Ryan \textbf{cost} 2 cents to everyone\\ Lily \textbf{wished} good health to him\\ You \textbf{wished} a good weekend to us\end{tabular}} \\ \cmidrule(l){2-5} 
 &
  PO &
  {\begin{tabular}[c]{@{}l@{}}address (1), announce (1), carry (31), \\ describe (4), drop (2), explain (20),\\ introduce (14), lift (2), mention (2), \\ return(5) , say (78), whisper (3)\end{tabular}} &
  {\begin{tabular}[c]{@{}l@{}}She \textbf{addressed} me that\\ You \textbf{carried} mommy some boxes\\ He \textbf{described} my uncle the day\\ I \textbf{explained} her everything\end{tabular}} &
  {\begin{tabular}[c]{@{}l@{}}She \textbf{addressed} that to me\\ You \textbf{carried} some boxes to mommy\\ He \textbf{described} the day to my uncle\\ I \textbf{explained} everything to her\end{tabular}} \\ \bottomrule
\end{tabular}%
}
\caption{\naba{} and \nana{} verb lemmas from AO-CHILDES along with test sentences in the two dative constructions. Values in parentheses following the verb lemmas denote frequency in the observed dative in the training data. The acceptability of the example test sentences should reflect the known alternation behavior---\naba{} alternates should be more acceptable, whereas \nana{} alternates should be less acceptable.}
\label{tab:nabananastats}
\end{table}

\section{Stimuli to evaluate LMs' alternation preferences on verbs that do not alternate in training}

An important precondition that we test our LM learners on is to be able to distinguish between two classes of verbs that do not alternate in the training data: 1) \naba{} verbs, which do not alternate in training but can actually alternate (Not Alternating in training data But actually Alternating); and 2) \nana{} verbs, which also do not alternate in training and do not tend to alternate (Not Alternating in training data and actually Not Alternating).
Using the method described above, we found a total of 6 NABA-DO verb lemmas, 2 NANA-DO verb lemmas, 6 NABA-PO verb lemmas, and 12 NANA-PO verb lemmas in the training data.
For each of these verbs, we manually constructed test sentences in both dative constructions with the following constraints placed on the event participants (agent, theme, recipient): (1) the lexical items should occur in the training corpus; (2) the event participants should obey the selectional preference of the verb (e.g., \textit{she kicked me the ball vs. \#she kicked the ball me}); and (3) the event participants should not appear linearly adjacent to the verb in the training set, to avoid the uninformative scenario where the co-occurrence statistics in the training data explain our results. For example, if the bigram \textit{you described} occurs in the corpus, then \textit{you} never appeared as an agent of \textit{described} in our test sentences.
For each verb, we selected 7 themes and 10 recipients, and then exhaustively created all possible theme-recipient combinations ($N=70$) with agents sampled from items in AO-CHILDES in accordance with the constraints above (i.e., occurrence of lexical item in training and no bigram overlap of the agent and verb with the training set). 
As a result, we ended up with 70 test sentences per verb, amounting to 840 \naba{} sentences ($N_{\text{PO}}$ = 420); and 960 \nana{} sentences ($N_{\text{PO}}$ = 840). \Cref{tab:nabananastats} shows the list of \naba{} and \nana{} verb lemmas in their observed dative constructions, as well as a few examples of the test sentences we constructed to analyze LM behavior on these verbs.


\section{Details of statistical analyses}

This section describes the details of all our statistical analyses. We use linear mixed-effects regression (LMER) as our main analysis technique throughout. The analysis were conducted in R (version 4.4.2), using the \texttt{lme4} \citep{lme4} and \texttt{lmerTest} \citep{lmerTest} libraries. 

To test if models show sensitivity to the alternation preferences of known verbs that do not alternate in training, we use the average log probability per token on the sentences in the dataset we constructed to compare \naba{} and \nana{} verbs (see Table 3 in Supplementary Materials for examples) as our dependent variable, and the classification (\naba{} vs. \nana{}) as our fixed effect, with random effects for the model seed.
To test if models show asymmetric cross-dative generalization \citep{conwell2007early}, we use the average log probability of the sentences belonging to the unmodeled form (i.e., PO if the exposure was DO) as the dependent variable, with the exposure dative (DO vs. PO) as the fixed effect, with random slopes for the model seed, and the discourse context template.
For the experiment pertaining to the advantage of cross structure training for the acquisition of DO datives  \citep{arunachalam2017preschoolers}, we use the average log probability per token of the DO sentences in the generalization set as the dependent variable, with the exposure dative (DO vs. PO) as the fixed effect, with random slopes per model seed.
To test if theme animacy has any effect in generalizing the verb from DO exposures with certain lexical items to other DO instances, we used the same dependent variable as in the previous analysis (average log probabilities of DO sentences in the generalization set), while subsetting the data only to DO exposures. We use the theme animacy as the fixed effect, and include random slopes for the model seed.

For our main simulations, we run two types of LMERs separately for the two experiments (DO to PO vs. PO to DO; 4 models total): one to compare the fit of our \haap{} scores vs. that of the counterfactual coding methods, and the second to further explore the role of theme and recipient arguments within \haap{} scoring. In both cases, we use as our dependent variable the average log probability of the generalization sentences belonging to the alternate, unmodeled construction (DO if exposure is PO and vice versa). For the first analysis, we use the score obtained under each scoring method as a fixed effect, with random effects for model seed, discourse template, and the item nested within each feature configuration. We then repeat this for each scoring technique, and compare the fit of the models to a null model (without the fixed effect) using the difference in log-likelihoods, where greater the difference, better is the model's fit.
Then for the second analysis, we focused only on the \haap{} scoring technique, since it yielded the best fit for both types of exposure dative constructions. Here, we decomposed the score into its constituents: sum of the binary scores for theme, sum of the binary scores for recipient, and the length score, and then used them as separate fixed effects, with the same random effects as before. 

\section*{Supplemental Note 1}
The exposure conditions are coded using 4 binary features for theme, 4 binary features for recipient, and the difference in the length of the first and second postverbal argument of the exposure ($\Delta\text{Length}$). We keep $\Delta\text{Length}$ as constant, and consider all plausible ways to code the 8 binary features. This gives us a total of 2\textsuperscript{8} = 256 codes. However, since each code has a unique complement, half of the codes are redundant. This is because we sum the codes (along with $\Delta\text{Length}$) to get the final code-score for our analysis, and the score obtained by complements of a given binary vector will strictly mirror the original sum (i.e., if a vector's features sum to $k$, its complement sums to $8-k$). As a result, a code and its complement will yield identical model fits, leaving exactly 256/2 = 128 mathematically distinct coding schemes to evaluate.



\end{document}